\documentclass[12pt]{iopart}
\usepackage{iopams, bm}
\usepackage{amssymb}
\usepackage{color}
\usepackage{graphicx}

\usepackage{pseudocode}
 \expandafter\let\csname \endcsname\relax
\renewcommand{\thepseudocode}{1}

\begin{document}

\title[Estimator of Prediction Error Based on AMP]
{Estimator of Prediction Error Based on Approximate Message Passing
for Penalized Linear Regression}

\author{Ayaka Sakata$^{1,2}$}

\address{$^1$The Institute of Statistical Mathematics, Midori-cho, Tachikawa, Tokyo 190-8562, Japan}
\address{$^2$The Graduate University for Advanced Science (SOKENDAI), Hayama-cho, Kanagawa 240-0193, Japan}
\ead{$^{1,2}$ayaka@ism.ac.jp}
\vspace{10pt}

\begin{abstract}
We propose an estimator of prediction error using an approximate message passing (AMP) algorithm
that can be applied to a broad range of sparse penalties.
Following Stein's lemma, the estimator of the generalized degrees of freedom,
which is a key quantity for the construction of the estimator of the prediction error,
is calculated at the AMP fixed point.
The resulting form of the AMP-based estimator does not depend on the penalty function,
and its value can be further improved by considering the correlation between predictors.
The proposed estimator is asymptotically unbiased when the components of the predictors and response variables are independently generated according to a Gaussian distribution.
We examine the behavior of the estimator for real data under nonconvex sparse penalties, where Akaike's information criterion does not correspond to an unbiased estimator of the prediction error.
The model selected by the proposed estimator is close to that which minimizes the true prediction error.
\end{abstract}

%
%
%

\section{Introduction}

In recent decades, variable selection using sparse penalties, referred to here as sparse estimation,
has become an attractive estimation scheme
\cite{lasso,Candes-Tao,Donoho}.
The sparse estimation is mathematically formulated as the
minimization of the estimating function associated with the sparse penalties.
In this paper, we concentrate on the linear regression 
problem with an arbitrary sparse regularization.
Let $\bm{y}\in\mathbb{R}^M$ and $\bm{A}\in\mathbb{R}^{M\times N}$ 
be a response vector and predictor matrix, respectively,
where each column of $\bm{A}$ corresponds to a predictor.
We denote the regression coefficient to be estimated as $\bm{x}\in\mathbb{R}^N$, 
and the problem considered here can be formulated as 
\begin{equation}
\min_{\bm{x}}\frac{1}{2}||\bm{y}-\bm{A}\bm{x}||_2^2+J(\bm{x};\eta),
\label{eq:original}
\end{equation}
where $J(\bm{x};\eta)=\sum_{i=1}^NJ(x_i;\eta)$ is the sparse regularization that enhances the zero component in $\bm{x}$, and $\eta$ is a set of regularization parameters.
In the classical variable selection problem,
the variables are estimated by a maximum likelihood method
under the constraint on the support of the variables used in the model.
In sparse estimation, the support is determined by the 
regularization parameter $\eta$; hence,
the adjustment of $\eta$ is regarded as the selection of a model, and 
is crucial in the sparse estimation.

In general, the regularization parameter is determined based on the model selection criteria.
One of the criteria is the prediction performance,
which is measured by the prediction error defined for the test data generated according to the true distribution.
Following this criterion, the model with the regularization parameter that minimizes the prediction error provides an appropriate description of the data.
However, the exact computation of the prediction error is impossible, because
we do not know the true distribution.
Therefore, the model selection is generally implemented using estimators of the prediction error, rather than the prediction error itself. For the maximum likelihood estimation, Akaike's information criterion (AIC) is an unbiased estimator of the prediction error
when the true generative model is included in the set of models under consideration, and is widely used for model selection when concentrating on the prediction of unknown data \cite{AIC}. 
However, for sparse estimation, AIC is generally not an unbiased estimator of the prediction error. 
In such cases, a $C_p$-type criterion, which evaluates the prediction error using generalized degrees of freedom (GDF) \cite{Mallows,Efron}, is useful when the variance of the data is known.
However, the exact computation of GDF also requires the true probability distribution of the data, which means that an estimator must be constructed.
In sparse estimation, GDF and its unbiased estimator can be derived for certain regularizations.
A well-known result obtained using the least absolute shrinkage and selection operator (LASSO) is that AIC corresponds to the unbiased estimator of the prediction error  \cite{GDF}, a nontrivial fact that has attracted the interest of statisticians.
Recently, sparse estimations using nonconvex regularizations have been studied with the aim of achieving a more compact representation of the data than LASSO \cite{SCAD,MCP}. Hence, model selection methods that can be applied to such regularizations are required. 

In this paper, we propose a numerical method for the construction of estimators for GDF and prediction error based on approximate message passing (AMP).
The derived form of GDF does not depend on the regularization, although the property of the AMP fixed point is regularization-dependent.
The estimator is asymptotically unbiased when the predictor matrix and response variables are independently and identically distributed (i.i.d.) Gaussian, which is indicated by showing the correspondence between AMP and the replica method \cite{Sakata_GDF}.
In this case, we can confirm that AIC is the unbiased estimator for the $\ell_1$ penalty, which is consistent with an earlier study \cite{GDF}.
We apply our estimator to real data and show that it adequately emulates model selection according to the prediction error.

The remainder of this paper is organized as follows.
In Sec.~\ref{sec:backgrounds}, we summarize the background materials that support the following discussion.
The problem setting considered in this paper is explained in Sec.~\ref{sec:problem_setting},
and the AMP algorithm for the penalized linear regression problem is introduced in 
Sec. \ref{sec:AMP}.
The estimator of GDF using the AMP fixed point is derived in Sec.~\ref{sec:AMP_estimator},
and its asymptotic property for Gaussian random data and predictor are studied in Sec.~\ref{sec:asymptotic} using the replica method. 
In Sec.~\ref{sec:application}, the proposed estimator is applied to real data, and 
the improvement it offers is quantified in Sec.~\ref{sec:correction}.
Finally, Sec.~\ref{sec:summary} presents the conclusions to this study and a discussion of the results.

\section{Background}
\label{sec:backgrounds}

\subsection{Prediction error and generalized degrees of freedom}

We denote the estimated regression coefficient obtained by solving \eref{eq:original} and its fit to the response variable
$\bm{y}$ as $\hat{\bm{x}}(\bm{y})$ and $\hat{\bm{y}}=\bm{A}\bm{\hat{x}}(\bm{y})$, respectively.
The prediction performance of the model 
under the given response variable is measured by
the prediction error 
\begin{equation}
\epsilon_{\rm pre}(\bm{y})
=\frac{1}{M}E_{\bm{z}}[||\bm{z}-\hat{\bm{y}}(\bm{y})||_2^2],
\label{eq:pre_err}
\end{equation}
where $\bm{z}\in\mathbb{R}^M$ represents test data whose
statistical property is equivalent to that of $\bm{y}$.
The value of $\eta$ that minimizes the prediction error gives an 
appropriate description of the data $\bm{y}$.
We aim to approximate the prediction error by constructing its estimator. 
The simplest estimator is the training error, which is
defined by
\begin{equation}
\epsilon_{\mathrm{train}}(\bm{y})=\frac{1}{M}||\bm{y}-\hat{\bm{y}}(\bm{y})||_2^2,
\end{equation}
but this underestimates the prediction error because of the correlation between $\bm{y}$ and
$\hat{\bm{y}}$.
We consider the case in which the mean and covariance matrix of the response vector $\bm{y}$ are given by
\begin{equation}
E_{\bm{y}}[\bm{y}]=\bm{m}_y,~~~E_{\bm{y}}[(\bm{y}-\bm{m}_y)(\bm{y}-\bm{m}_y)^{\rm T}]=\sigma_y^2\bm{I}_M,
\label{eq:y_gauss}
\end{equation}
where $\mathrm{T}$ denotes the matrix transpose, and $\bm{I}_M$ is the $M$-dimensional identity matrix.
In this case, the prediction error and the training error satisfy the relationship
\begin{equation}
E_{\bm{y}}[\epsilon_{\mathrm{pre}}(\bm{y})]=E_{\bm{y}}[\epsilon_{\rm train}(\bm{y})]+2\sigma^2_y\mathrm{df},
\label{eq:c_p}
\end{equation}
where $\mathrm{df}$ is the {\it generalized degrees of freedom}
(GDF) defined by \cite{Ye1998} 
\begin{equation}
{\rm df}=\frac{{\rm cov}(\bm{y},\hat{\bm{y}}(\bm{y}))}{M\sigma^2},
\label{eq:df_def}
\end{equation}
and ${\rm cov}(\bm{y},\hat{\bm{y}})=E_{\bm{y}}[\bm{y}^{\rm T}\hat{\bm{y}}]-E_{\bm{y}}[\bm{y}]^{\rm T}E_{\bm{y}}[\hat{\bm{y}}]$.
The expression \eref{eq:c_p} is known as the $C_p$ criterion for model selection~\cite{Mallows,Efron}.

\subsection{Stein's lemma}

The covariance \eref{eq:df_def} is not observable 
in the sense that the expectation with respect to the true distribution is required.
Hence, estimators of df, denoted by $\hat{\mathrm{df}}(\bm{y})$, are used instead of $\mathrm{df}$ for approximating the prediction error.
An idea for the construction of the estimator appears in 
Stein's unbiased risk estimate (SURE), known as Stein's lemma \cite{Stein}.
Following Stein's lemma,
the component-wise covariance between the response variable and its fit is given by
\begin{equation}
\mathrm{cov}[y_\mu,\hat{y}_\mu(\bm{y})]=\sigma_y^2E_{\bm{y}}\left[\frac{\partial \hat{y}_\mu(\bm{y})}{\partial y_\mu}\right].
\label{eq:stein}
\end{equation}
\Eref{eq:stein} means that 
an unbiased estimator of GDF is given by 
\begin{equation}
\hat{\mathrm{df}}(\bm{y})=\frac{1}{M}\sum_{\mu=1}^M\frac{\partial \hat{y}_\mu(\bm{y})}{\partial y_\mu}.
\label{eq:df_divergence}
\end{equation}
Note that \eref{eq:stein} is exact when the response variables are i.i.d. Gaussian random variables.\Eref{eq:df_divergence} is an observable quantity, and hence one of the estimators of the prediction error can be defined by
\begin{equation}
\hat{\epsilon}_{\mathrm{pre}}(\bm{y})=\mathrm{\epsilon}_{\mathrm{train}}(\bm{y})+\frac{2\sigma_y^2}{M}\sum_{\mu=1}^M\frac{\partial \hat{y}_\mu(\bm{y})}{\partial y_\mu}.
\label{eq:e_pre_estimator}
\end{equation}
\Eref{eq:e_pre_estimator} is an unbiased estimator of the prediction error when the response variables are i.i.d. Gaussian, but the unbiasedness is not generally guaranteed
for other response variables. 

For the maximum likelihood method,
the unbiased estimator of GDF is given by the number of parameters used in the model, which is the AIC given by \cite{AIC}
\begin{equation}
\mathrm{AIC}(\bm{y})=\epsilon_{\mathrm{train}}(\bm{y})+\frac{2}{M}\sigma_y^2||\hat{\bm{x}}(\bm{y})||_0,
\end{equation}
when the true generative model is included in the assumed models.
As defined here, AIC has been multiplied by $\sigma_y^2\slash M$ compared with its conventional definition, but this does not affect the following discussion.
For penalized regression problems, when the fit of the response variable is represented as $\hat{\bm{y}}=\bm{H}\bm{y}$
using a matrix $H$, it can be shown that GDF corresponds to $\mathrm{Tr}\bm{H}$ \cite{GDF_H},
which is widely used in ridge regression.
Furthermore, many studies have derived SURE in problems where \eref{eq:df_divergence} is easily computed \cite{Kneip, Li1987, Donoho_Johnstone}.

\subsection{GDF for sparse penalties}

For sparse penalties, the derivation of GDF and the construction of its estimator is not straightforward, because the penalty term is not differentiable at the origin.
For the $\ell_1$ penalty, the GDF is equal to the expectation of the density of non-zero components \cite{GDF};
\begin{equation}
\mathrm{df} = \frac{1}{M}E_{\bm{y}}[||\hat{\bm{x}}(\bm{y})||_0],
\end{equation}
and hence 
$||\hat{\bm{x}}(\bm{y})||_0\slash M$ is an unbiased estimator of GDF.
It means that AIC corresponds to an unbiased estimator of the prediction error.
The extension of this result for general problem settings has been extensively discussed
\cite{Tibshirani-Taylor1, Tibshirani-Taylor2}.
However, such a simple relationship does not generally hold, 
and efficient computation methods for \eref{eq:df_divergence} are required.
Computation techniques for df have been developed using cross-validation and the parametric bootstrap method, but these techniques are computationally expensive \cite{Efron,Shen_Ye}.


In this paper, we propose a method to construct the estimator of the prediction error based on Stein's lemma using AMP. In principle, the method can be applied for any $J(x;\eta)$, and 
the extra computational cost of obtaining the fixed points is less than in other methods.   

\subsection{Sparse penalties considered in this paper}

\begin{figure}
\centering
\includegraphics[width=1.8in]{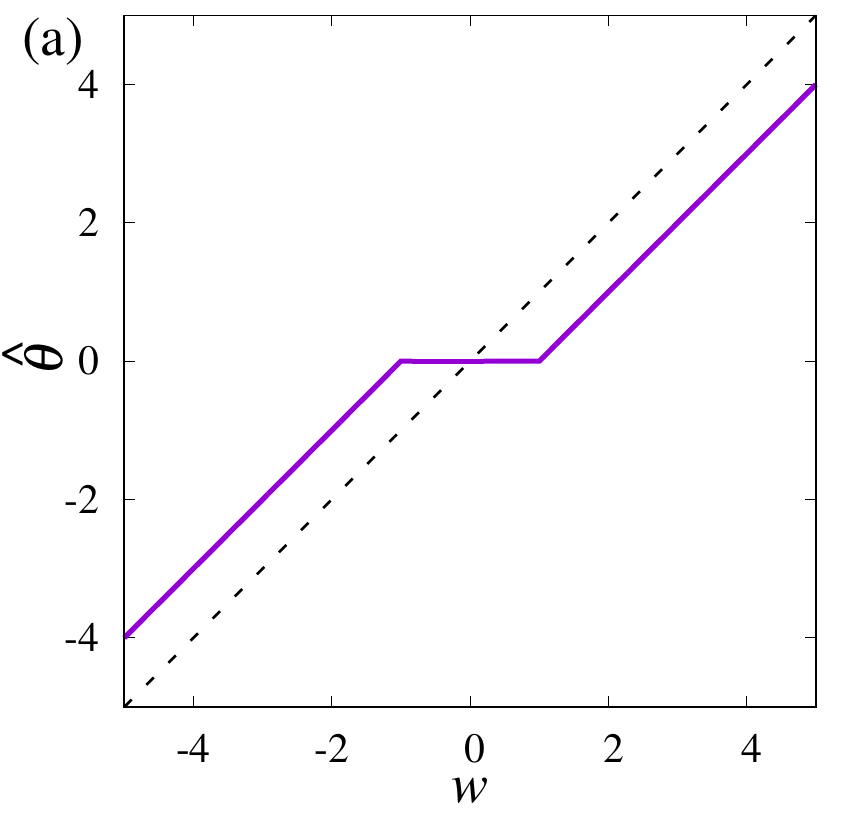}
\includegraphics[width=1.8in]{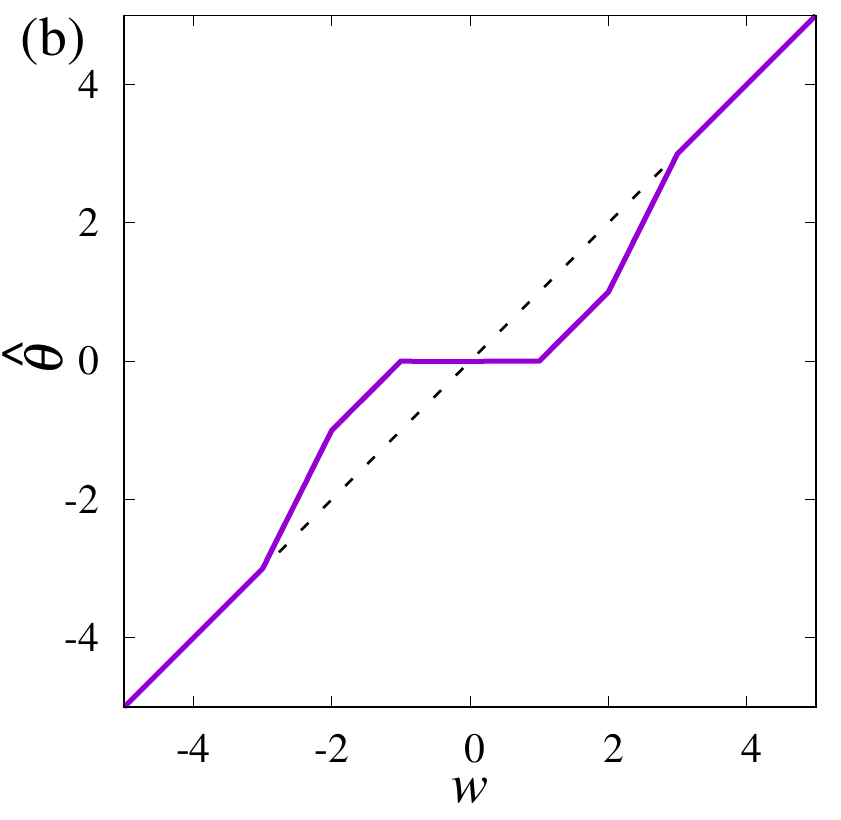}
\includegraphics[width=1.8in]{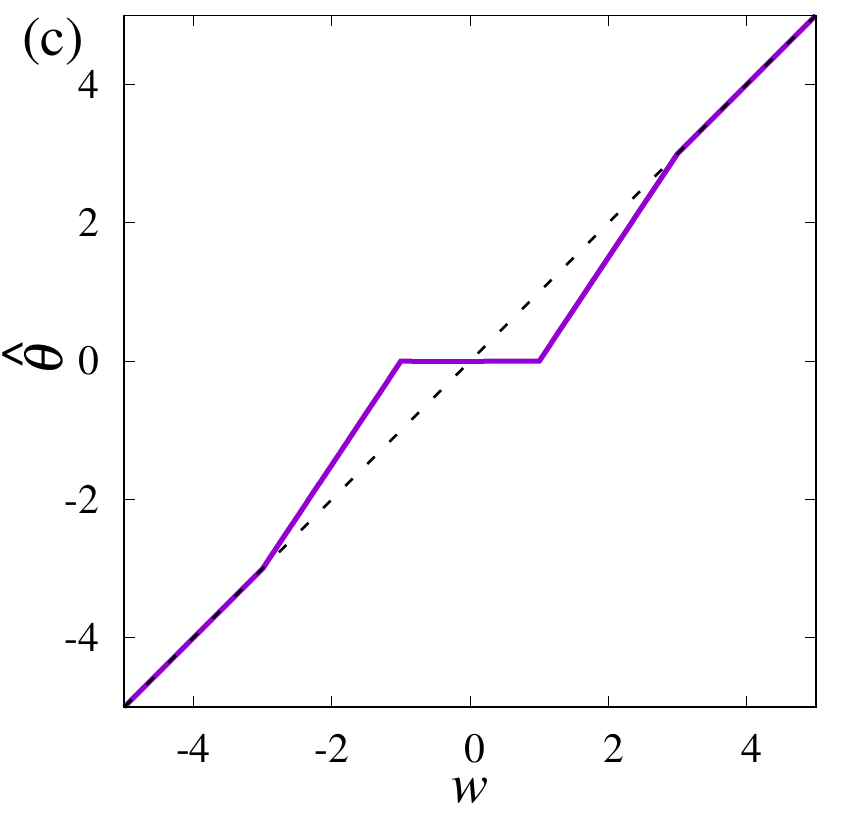}
\caption{Behavior of the estimator for (a) $\ell_1$, (b) SCAD, and (c) MCP.}
\label{fig:estimators}
\end{figure}

The penalty functions considered in this paper are $\ell_1$, the smoothly clipped absolute deviation (SCAD), and the minimax concave penalty (MCP).
Under these regularizations,
we explain the properties of the estimator using the one-dimensional problem
\begin{equation}
\hat{\theta}(w)=\arg\min_\theta\left\{\frac{1}{2\sigma_w^2}(\theta-w)^2+J(\theta;\eta)\right\},
\label{eq:one_dim}
\end{equation}
where $w$ is the training sample.
The ordinary least square (OLS) estimator, which minimizes the squared error between $\theta$ and $w$, is $\hat{\theta}_{\mathrm{OLS}}(w) = w$ for any $w$.
The penalty term changes the behavior of the estimator from that of the OLS estimator. 
The penalties considered here give analytical solutions of \eref{eq:one_dim}.
As explained in Sec.~\ref{sec:AMP},
the estimates given by AMP reduce to the one-body problem \eref{eq:one_dim}
with effective $\sigma_w^2$ and $w$.
We summarize the solution of \eref{eq:one_dim} in the following subsections.

\subsubsection{$\ell_1$ penalty}

The $\ell_1$ penalty is given by
\begin{equation}
J(\theta;\lambda)=\lambda|\theta|,
\end{equation}
where $\lambda$ is the regularization parameter determined by the 
model selection criterion.
The solution of \eref{eq:one_dim} under the $\ell_1$ penalty is given by
\begin{equation}
\hat{\theta}(w)=V_{\ell_1}(\tilde{w};\sigma_w^2,\lambda)S_{\ell_1}(\tilde{w};\sigma_w^2,\lambda)
\end{equation}
where $\tilde{w}=w\slash\sigma_w^2$ and 
\begin{eqnarray}
S_{\ell_1}(w;\sigma^2,\lambda)&=\left\{\begin{array}{ll}
w-\mathrm{sgn}(w)\lambda & \mathrm{for}~|w|>\lambda\\
0 & \mathrm{otherwise}
\end{array}
\right.\\
V_{\ell_1}(w;\sigma^2,\lambda)&=\left\{\begin{array}{ll}
\sigma^2 & \mathrm{for}~|w|>\lambda\\
0 & \mathrm{otherwise}
\end{array}
\right..
\end{eqnarray}
The behavior of the estimator at $\lambda=1$ is shown in \Fref{fig:estimators}(a).
Across the whole region of $w$,
the $\ell_1$ estimator is shrunk from the ordinary least square (OLS) estimator,
denoted by the dashed line.

\subsubsection{SCAD penalty}
\label{sec:one_body_SCAD}

The SCAD penalty is defined by \cite{SCAD}
\begin{equation}
J(\theta;\eta)=\left\{\begin{array}{ll}
\lambda|\theta| & (|\theta|\leq \lambda) \\
-\displaystyle\frac{\theta^2-2a\lambda|\theta|+\lambda^2}{2(a-1)} & (\lambda<|\theta|\leq a\lambda) \\
\displaystyle\frac{(a+1)\lambda^2}{2} & (|\theta|>a\lambda)
\end{array}
\right.,
\end{equation}
where the regularization parameter is $\eta=\{\lambda,a\}$.
The solution of \eref{eq:one_dim} under SCAD regularization is given by
\begin{equation}
\hat{\theta}(w;\sigma_w^2,\eta)=V_{\mathrm{SCAD}}(\tilde{w};\sigma_w^2,\eta)S_{\mathrm{SCAD}}(\tilde{w};\sigma_w^2,\eta),
\end{equation}
where
\begin{eqnarray}
S_{\mathrm{SCAD}}(w;\sigma^2,\eta)&=\left\{\begin{array}{ll}
w-\mathrm{sgn}(w)\lambda & \mathrm{for}~\lambda(1+\sigma^{-2})\geq|w|>\lambda\\
w-\mathrm{sgn}(w)\frac{a\lambda}{a-1} & \mathrm{for}~a\lambda\sigma^{-2}\geq|w|>\lambda(1+\sigma^{-2})\\
w & \mathrm{for}~|w|>a\lambda\sigma^{-2}\\
0 & \mathrm{otherwise}
\end{array}
\right.\\
V_{\mathrm{SCAD}}(w;\sigma^2,\eta)&=\left\{\begin{array}{ll}
\sigma^2 & \mathrm{for}~\lambda(1+\sigma^{-2})\geq|w|>\lambda\\
\left(\sigma^{-2}-\frac{1}{a-1}\right)^{-1}& \mathrm{for}~a\lambda\sigma^{-2}\geq|w|>\lambda(1+\sigma^{-2})\\
\sigma^2 & \mathrm{for}~|w|>a\lambda\sigma^{-2}\\
0 & \mathrm{otherwise}
\end{array}
\right.
\end{eqnarray}
The behavior of the estimator under SCAD regularization at $a=3$ is shown in \Fref{fig:estimators}(b).
The SCAD estimator behaves like the $\ell_1$ and OLS estimators when 
$\lambda(1+\sigma^{-2})\geq|z|>\lambda$ and $|w|>a\lambda\sigma^{-2}$, respectively. In the region $a\lambda\sigma^{-2}\geq|w|>\lambda(1+\sigma^{-2})$,
the estimator linearly transits between the $\ell_1$ and OLS estimators.
We call this region the transient region of SCAD.

\subsubsection{MCP}
\label{sec:one_body_MCP}

The MCP is defined by \cite{MCP}
\begin{eqnarray}
J(\theta;\eta)&=\left\{\begin{array}{ll}
\lambda|\theta|-\displaystyle\frac{\theta^2}{2a} & (|\theta|\leq a\lambda) \\
\displaystyle\frac{a\lambda^2}{2} & (|\theta|> a\lambda) 
\end{array}
\right.,
\end{eqnarray}
where $\eta=\{\lambda,a\}$. The estimator of \eref{eq:one_dim} under MCP is given by
\begin{equation}
\hat{\theta}(w;\sigma_w^2,\eta)=V_{\mathrm{MCP}}(\tilde{w};\sigma_w^2,\eta)S_{\mathrm{MCP}}(\tilde{w};\sigma_w^2,\eta),
\end{equation}
where
\begin{eqnarray}
S_{\mathrm{MCP}}(w;\sigma^2,\eta)&=\left\{\begin{array}{ll}
w-\mathrm{sgn}(w)\lambda & \mathrm{for}~a\lambda\sigma^{-2}\geq|w|>\lambda\\
w & \mathrm{for}~|w|>a\lambda\sigma^{-2}\\
0 & \mathrm{otherwise}
\end{array}
\right.\\
V_{\mathrm{MCP}}(w;\sigma^2,\eta)&=\left\{\begin{array}{ll}
(\sigma^{-2}-a^{-1})^{-1} & \mathrm{for}~a\lambda\sigma^{-2}\geq|w|>\lambda\\
\sigma^{-2} & \mathrm{for}~|w|>a\lambda\sigma^{-2}\\
0 & \mathrm{otherwise}
\end{array}
\right.
\end{eqnarray}
\Fref{fig:estimators}(c) shows the behavior of the MCP estimator at $a=3$.
The MCP estimator behaves like the OLS estimator when $|w|>a\lambda\sigma^{-2}$,
and is connected from zero to the OLS estimator in the region $a\lambda\sigma^{-2}\geq|w|>\lambda$. We call this the transient region of MCP.

\section{Problem setting}
\label{sec:problem_setting}

Problem \eref{eq:original} discussed in this paper is equivalent to searching the
ground state,
but here we extend the problem to finite temperatures
by introducing the inverse temperature $\beta$.
We construct the posterior distribution for finite $\beta$,
and extract the ground state corresponding to the
maximum a posteriori (MAP) estimator
in the limit $\beta\to\infty$.

The likelihood we consider here is the Gaussian distribution given by
\begin{equation}
P_l(\bm{y}|\bm{x})\propto\exp\left\{-\frac{\beta}{2}||\bm{y}-\bm{Ax}||_2^2\right\}.
\label{eq:likelihood}
\end{equation}
The distribution \eref{eq:likelihood} depends on the parameter $\bm{x}$ only through the form of $\bm{Ax}$; hence, we hereafter denote \eref{eq:likelihood} as $P_l(\bm{y}|\bm{u}(\bm{A},\bm{x}))$,
where $\bm{u}(\bm{A},\bm{x})=\bm{Ax}$.
The prior distribution of $\bm{x}$ is given by the penalty function and the inverse temperature as 
\begin{equation}
P_r(\bm{x})\propto\exp(-\beta J(\bm{x})),
\end{equation}
and hence the posterior distribution of $\bm{x}$
is given by
\begin{equation}
P_\beta(\bm{x}|\bm{y})=\exp\left\{-\frac{\beta}{2}||\bm{y}-\bm{Ax}||_2^2-\beta J(\bm{x}) -\ln Z_\beta(\bm{y})\right\}
\label{eq:posterior}
\end{equation}
following Bayes' formula,
where $Z_\beta(\bm{y})$ is the normalization constant.
The estimate for the solution of \eref{eq:original} is expressed as 
\begin{equation}
\hat{\bm{x}}(\bm{y})=\lim_{\beta\to\infty}\langle\bm{x}\rangle_\beta,
\label{eq:def_estimate}
\end{equation} 
where $\langle\cdot\rangle_\beta$ denotes the expectation according to \eref{eq:posterior}
at $\beta$. As $\beta\to\infty$, \eref{eq:posterior} becomes the uniform distribution 
over the minimizers of \eref{eq:original}. Hence,
when problem \eref{eq:original} has a unique minimizer,
it corresponds to the estimate \eref{eq:def_estimate}.

The exact computation of the expectation in \eref{eq:def_estimate}
is intractable.
Thus, we resort to the approximated message passing (AMP) algorithm to approximate \eref{eq:def_estimate}, 
a technique that is widely used in signal processing and statistical physics
\cite{Mezard-Montanari,CDMA,AMP,GAMP}.

\section{Approximate message passing}
\label{sec:AMP}

In this section, we briefly summarize AMP for penalized regression problems. The explanation here only covers the aspects required to reach the result shown in Sec.~\ref{sec:AMP_estimator}; a detailed derivation is given in \cite{Sakata_Xu}.
In the derivation of AMP, we introduce the following assumptions
for sufficiently large $M$ and $N$.

\begin{description}

\item[A1:] All components of the predictor matrix are $O(M^{-1\slash 2})$.

\item[A2:] The correlation between predictors is negligible: the off-diagonal components of 
$\bm{A}^{\mathrm{T}}\bm{A}$ are $O(M^{-1\slash 2})$.

\end{description}

The message passing algorithm defined for the probability distribution $P(\bm{x}|\bm{y},\bm{A})$ is
expressed by $2MN$ \textit{messages} that are propagated
from factors to variables $\hat{m}_{\mu\to i}$ and 
variables to factors $m_{i\to\mu}$ according to
\begin{eqnarray}
\hat{m}_{\mu\to i}(x_i)&=&\frac{1}{\hat{Z}_{\mu\to i}}\int \prod_{j\in{\cal V}(\mu)\backslash i}dx_j P_{l}(y_\mu|u_\mu(\bm{A},\bm{x}))\prod_{j\in{\cal V}(\mu)\backslash i}m_{j\to\mu}(x_j)\label{eq:hat_message}\\
m_{i\to\mu}(x_i)&=&\frac{1}{Z_{i\to\mu}}P_{r}(x_i)\prod_{\gamma\in{\cal F}(i)\backslash\mu}\hat{m}_{\gamma\to i}(x_i).
\label{eq:message}
\end{eqnarray}
Here, ${\cal V}(\mu)$ and ${\cal F}(i)$ represent the variable nodes and factor nodes connected to the $\mu$-th factor node and $i$-th variable node, respectively,
and $\backslash i$ denotes that $i$ is not included.
Using the messages, the marginal distribution is given by
\begin{equation}
m_i(x_i)=\frac{1}{Z_i}P_{r}(x_i)\prod_{\gamma\in{\cal M}(i)}\hat{m}_{\gamma\to i}(x_i).
\end{equation}

Updating the messages in the form of the probability distribution is intractable, and we 
represent the distributions using the mean and rescaled variance,
which is the variance multiplied by $\beta$, assuming that they can be defined:
\begin{eqnarray}
a_{i\to\mu}&\equiv&\int dx x m_{i\to\mu}(x)\\
v_{i\to\mu}&\equiv&\beta\int dx (x-a_{i\to\mu})^2 m_{i\to\mu}(x).
\end{eqnarray}
Further, we define the mean and rescaled variance with respect to the marginal distribution as
\begin{eqnarray}
a_i&\equiv&\int dx xm_i(x)\\
v_i&\equiv&\beta\int dx(x-a_i)^2m_i(x),
\label{eq:vi_def}
\end{eqnarray}
where the $i$-th component of the estimate \eref{eq:def_estimate}, denoted by $\hat{x}_i$, corresponds to $a_i$ at $\beta\to\infty$.
Following the calculation under assumptions A1 and A2, 
the 
marginal distribution is derived as \cite{Sakata_Xu}
\begin{equation}
m_i(x)={\cal M}(x;\Sigma_i^2,R_i),
\label{eq:marginal_dist}
\end{equation}
where 
\begin{eqnarray}
&\Sigma^{2}_i=\left(\sum_{\mu}A_{\mu i}^2(g_{\rm out}^{\prime})_{\mu} \right)^{-1}, \label{eq:Sigma}\\
&R_i=a_{i}+\left( \sum_{\mu}(g_{\rm out})_{\mu}A_{\mu i}\right)\Sigma^{2}_i,\label{eq:R}\\
&{\cal M}(x;\Sigma^2,R)=\frac{1}{\hat{Z}(\Sigma^2,R)}P_r(x)\frac{1}{\sqrt{2\pi\Sigma^2}}
\exp\left\{-\frac{(x-R)^2}{2\Sigma^2}\right\},
\end{eqnarray}
and $\hat{Z}(\Sigma^2_i,R_i)$ is the normalization constant required to satisfy
$\int dx{\cal M}(x;\Sigma^2,R)=1$.
Equation (\ref{eq:marginal_dist}) suggests that the product of messages 
from factors connected to $i$, $\prod_{\nu\in{\cal F}(i)}\hat{m}(y_\nu)$,
corresponds to a Gaussian distribution with mean $R_i$ and variance $\Sigma^2_i$.
The functions $g_{\mathrm{out}}$ and $g_{\mathrm{out}}^\prime$ are given by 
\begin{eqnarray}
(g_{\mathrm{out}})_\mu&=&\frac{\partial}{\partial\omega_\mu}\log\Xi(y_\mu|\omega_\mu,V_\mu)
\label{eq:g_out}\\
(g_{\mathrm{out}}^\prime)_\mu&=&-\frac{\partial^2}{\partial\omega_\mu^2}\log\Xi(y_\mu|\omega_\mu,V_\mu),
\label{eq:g_out_p}
\end{eqnarray}
where we define
\begin{eqnarray}
\omega_\mu&\equiv&\sum_{i=1}^N A_{\mu i}a_{i\to\mu}\label{eq:omega_def}\\
V_\mu&\equiv&\sum_{i=1}^NA_{\mu i}^2v_i,
\end{eqnarray}
and the function $\Xi(y_\mu|\omega_\mu,V_\mu)$ contains the dependence on the likelihood as 
\begin{equation}
\Xi(y_\mu|\omega_\mu,V_\mu)=\int du_\mu P_l(y_\mu|u_\mu)\exp\left\{-\frac{\beta(u_\mu-\omega_\mu)^2}{2V_\mu}\right\}.
\end{equation}
Under assumption A1, the following relationship holds \cite{Sakata_Xu}:
\begin{equation}
a_{i\to\mu}=a_i-v_i\left((g_{\mathrm{out}})_\mu A_{\mu i}+(g_{\mathrm{out}}^\prime)_\mu A_{\mu i}^2a_{i\to\mu}\right),
\end{equation}
which leads to
\begin{equation}
\omega_\mu=\sum_i A_{\mu i}a_i-\beta^{-1}(g_{\rm out})_\mu V_\mu.
\end{equation}

\subsection{Specific form for penalized linear regression}

Substituting the output function for linear regression given by
\begin{equation}
P_l(y_\mu|u_\mu)=\exp\left\{-\frac{\beta}{2}(y_\mu-u_\mu)^2\right\}
\label{eq:P_output}
\end{equation}
into Eqs.~(\ref{eq:g_out}) and (\ref{eq:g_out_p}),
we obtain the specific form of $g_{\mathrm{out}}$ and $g^\prime_{\mathrm{out}}$ as
\begin{eqnarray}
(g_{\rm out})_\mu&=&\frac{y_\mu-\omega_\mu}{1+V_\mu}\\
(g_{\rm out}^\prime)_\mu &=& \frac{1}{1+V_\mu}.
\end{eqnarray}
Using these equations, the mean and variance parameters are given by 
\begin{eqnarray}
R_i&=&a_i+\left(\sum_\mu \frac{y_\mu-\omega_\mu}{1+V_\mu} A_{\mu i}\right)\Sigma_i^2
\label{eq:R_linear}\\
\Sigma_i^2&=&\left(\sum_\mu \frac{A_{\mu i}^2}{1+V_\mu}\right)^{-1}.
\label{eq:Sigma_linear}
\end{eqnarray}
In summary, the AMP algorithm consists of \eref{eq:AMP_V}--\eref{eq:AMP_v_i}.
\begin{eqnarray}
V^{(t)}_\mu&=&\sum_iA_{\mu i}^2v_i^{(t-1)},\label{eq:AMP_V}\\
{\Sigma^2_i}^{(t)}&=&\left(\sum_\mu\frac{A_{\mu i}^2}{1+V^{(t)}_\mu}\right)^{-1}\\
\omega_\mu^{(t)}&=&\sum_{i=1}^NA_{\mu i}a_i^{(t-1)}-\frac{V_\mu^{(t)}(y_\mu-\omega_\mu^{(t-1)})}{1+V_\mu^{(t)}}\label{eq:omega}\\
R_i^{(t)}&=&a_i^{(t-1)}+\left(\sum_\mu \frac{y_\mu-\omega_\mu^{(t)}}{1+V_\mu^{(t)}} A_{\mu i}\right){\Sigma_i^2}^{(t)}\label{eq:R_final}\\
a_i^{(t)}&=&f_a({\Sigma_i^2}^{(t)},R_i^{(t)}),\\
v_i^{(t)}&=&f_c({\Sigma_i^2}^{(t)},R_i^{(t)})\label{eq:AMP_v_i},
\end{eqnarray}
where we define
\begin{eqnarray}
f_a(\Sigma^2,R)&=&\int dx x{\cal M}(x;\Sigma^2,R)\label{eq:def_f_a}\\
f_c(\Sigma^2,R)&=&\int dx (x-f_a(\Sigma^2,R))^2{\cal M}(x;\Sigma^2,R),
\end{eqnarray}
and note that
\begin{equation}
f_c(\Sigma_i^2,R_i)=\Sigma_i^2\frac{\partial f_a(\Sigma^2,R)}{\partial R_i}.
\label{eq:f_c_and_f_a}
\end{equation}
At $\beta\to\infty$,
the integral of \eref{eq:def_f_a} can be computed by the saddle point method as
\begin{equation}
f_a(\Sigma^2,R)=\arg\min_x\left\{\frac{(x-R)^2}{2\Sigma^2}-J(x)\right\},
\label{eq:f_a_one_body}
\end{equation}
which corresponds to the one-dimensional problem \eref{eq:one_dim}
with the replacements $\sigma_w^2\to\Sigma^2$ and 
$w\to R$.
Under the penalties considered here, namely, $\ell_1$, SCAD, and MCP penalties, these functions are computed analytically, and have the form 
\begin{eqnarray}
f_a(\Sigma^2,R)&=&V_{r}(\tilde{R};\Sigma^2,\eta)S_{r}(\tilde{R};\Sigma^2,\eta)\\
f_c(\Sigma^2,R)&=&V_{r}(\tilde{R};\Sigma^2,\eta),
\end{eqnarray}
where $\tilde{R}=R\slash\Sigma^2$
and the subscript $r$ denotes dependence on the regularization: $r\in\{\ell_1,\mathrm{SCAD},\mathrm{MCP}\}$.

\section{AMP-based estimator}
\label{sec:AMP_estimator}

At the AMP fixed point, Stein's lemma implies that the covariance between the response variable and its fit can be expressed as 
\begin{equation}
\mathrm{cov}[y_\mu,\hat{y}_\mu]=\sigma_y^2E_{\bm{y}}\left[\sum_\nu A_{\nu i}\frac{\partial a_i}{\partial y_\nu}\right].
\end{equation}
The dominant dependence on $y_\nu$ of the estimate $a_i$ is induced through the mean parameter $R_i$. Hence, using \eref{eq:f_c_and_f_a},
\begin{equation}
\frac{d a_i}{d y_\nu}\sim
\frac{d R_i}{d y_\nu}\frac{\partial f_a(\Sigma^2_i,R_i)}{\partial R_i}=\frac{v_i}{\Sigma_i^2}\frac{d R_i}{d y_\nu}.
\label{eq:a_derivative}
\end{equation}
Substituting the definition of $\omega_\mu$, \eref{eq:omega_def}, into \eref{eq:R_final},
we obtain
\begin{equation}
R_i=
\left(\sum_\mu\frac{y_\mu-\sum_{j\neq i}A_{\mu j}a_{j\to\mu}}{1+V_\mu}A_{\mu i}\right)\Sigma_i^2
+\left\{a_i-\sum_\mu\frac{A_{\mu i}^2a_{i\to\mu}}{1+V_\mu}\Sigma_i^2\right\},
\label{eq:R_proof}
\end{equation}
where the second term can be ignored under the assumption A1:
\begin{equation*}
a_i-\sum_\mu\frac{A_{\mu i}^2a_{i\to\mu}}{1+V_\mu}\Sigma_i^2=v_i\Sigma_i^2\sum_\mu\frac{A_{\mu i}^2}{1+V_\mu}\left((g_{\mathrm{out}})_\mu A_{\mu i}+(g_{\mathrm{out}}^\prime)_\mu A_{\mu i}^2a_{i\to\mu}\right)=O(M^{-1}).
\end{equation*}
Hence, we have 
\begin{eqnarray}
\frac{d R_j}{d y_\nu}&=&\Sigma_j^2\left\{\frac{A_{\nu j}}{1+V_\nu}-\sum_\kappa\frac{\sum_{k\neq j}A_{\kappa k}A_{\kappa j}\frac{\partial a_{k\to \kappa}}{\partial y_\nu}}{1+V_\kappa}\right\}\label{eq:dR_1}\\
&=& \Sigma_j^2\left\{\frac{A_{\nu j}}{1+V_\nu}-\sum_\kappa\frac{\sum_{k\neq j}A_{\kappa k}A_{\kappa j}\frac{\partial R_{k}}{\partial y_\nu}\frac{\partial f_a(\Sigma_k^2,R_k)}{\partial R_k}}{1+V_\kappa}\right\},\label{eq:dR_2}
\end{eqnarray}
where $a_{k\to\kappa}$ is approximated by $a_k$ and \eref{eq:a_derivative} is used to 
derive \eref{eq:dR_2} from \eref{eq:dR_1}.
Substituting \eref{eq:dR_2} into \eref{eq:a_derivative}, 
the second term of \eref{eq:dR_2} can be ignored, and we obtain
\begin{equation}
E_{\bm{y}}\left[\sum_i A_{\nu i}\frac{\partial a_i}{\partial y_\nu}\right] 
=\sigma_y^2\sum_{\nu=1}^ME_y\left[\frac{V_\nu}{1+V_\nu}\right].
\label{eq:cov_AMP}
\end{equation}
This means that 
\begin{equation}
\hat{\mathrm{df}}^{(1)}(\bm{y})\equiv\frac{1}{M}\sum_\nu\frac{V_\nu}{1+V_\nu}
\label{eq:df_AMP_1}
\end{equation}
is the AMP expression of the unbiased estimator of GDF,
and we define the corresponding estimator of the prediction error as
\begin{equation}
\hat{\epsilon}_{\mathrm{pre}}^{(1)}(\bm{y})=\epsilon_{\mathrm{train}}(\bm{y})+2\sigma_y^2\hat{\mathrm{df}}^{(1)}(\bm{y}).
\label{eq:pre_estimator_1}
\end{equation}
These expressions do not depend on the form of the function $J(x;\eta)$,
hence they can be applied to any regularization $J(x;\eta)$,
such as elastic-net.
In deriving \eref{eq:df_AMP_1}, note that assumptions A1 and A2 are required,
and hence its unbiasedness is not generally guaranteed.

\section{Asymptotic behavior of the AMP-based estimator for Gaussian predictor matrix}
\label{sec:asymptotic}

When the components of the predictor matrix and data are i.i.d. according to a Gaussian distribution,
the asymptotic property of the AMP fixed point can be described by the replica method,
where $N\to\infty$ and $M\to\infty$ while $\alpha=M\slash N$ remains $O(1)$.
Here, we concentrate on the case 
$y_\mu\sim{\cal N}(0,\sigma_y^2)$ and $A_{\mu i}\sim {\cal N}(0,M^{-1})$,
but the discussion can be extended to the case in which the generative model of $\bm{y}$ contains 
a ``true'' sparse expression $\bm{x}^0$ such as $\bm{y}=\bm{A}\bm{x}^0+\bm{\xi}$, where $\bm{\xi}\in\mathbb{R}^M$ is a noise vector. 
It is known that the Gaussian i.i.d. predictor matrix is a full-rank matrix with probability 1; the rank of $\bm{A}$ is equal to $\min(N,M)$.

\subsection{Replica analysis}

The basis of replica analysis is the derivation of the free energy density, which is defined by
\begin{equation}
f\equiv -\lim_{\beta\to\infty}\frac{1}{M\beta}E_{\bm{y},\bm{A}}[\ln Z_\beta(\bm{y},\bm{A})].
\end{equation}
The free energy density and physical quantities derived from it are averaged over the predictor matrix and response variable, a procedure that is less common in the context of statistics. 
This averaging is implemented with the purpose of describing the typical behavior of the problem under consideration.

After the calculation under the replica symmetry assumption, the 
free energy density is derived as \cite{Sakata_Xu}
\begin{equation}
f=\mathop{\rm extr}_{Q,\chi,\hat{Q},\hat{\chi}}\Big\{\frac{Q+\sigma_y^2+m_y^2}{2(1+\chi)}-\frac{Q\hat{Q}-\chi\hat{\chi}}{2}
+\frac{1}{2\alpha}\xi(\hat{Q},\hat{\chi})\Big\},
\label{eq:f_replica}
\end{equation}
where
${\rm extr}_{Q,\chi,\hat{Q},\hat{\chi}}$
denotes extremization with respect to the
variables $\{Q,\chi,\hat{Q},\hat{\chi}\}$.
The function $\xi$
 is given by
\begin{eqnarray}
\xi(\hat{Q},\hat{\chi})&=2\int Dz\log f_\xi(\sqrt{\hat{\chi}}z,\hat{Q})\label{eq:f_pi_def}
\\
f_\xi(\sqrt{\hat{\chi}}z,\hat{Q})&=\min_x\exp\Big(\frac{\hat{Q}}{2}x^2-\sqrt{\hat{\chi}}zx+J(x;\eta)\Big),\label{eq:one_body}
\end{eqnarray}
where $\sqrt{\hat{\chi}}z$
is the random field that effectively represents the
randomness of the problem introduced by $\bm{y}$
and $\bm{A}$,
and $Dz=dz\exp(-z^2\slash 2)\slash\sqrt{2\pi}$.
\Eref{eq:one_body} is equivalent to the one-dimensional problem \eref{eq:one_dim}
with the correspondence $\sigma_w^2\to\hat{Q}^{-1}$ and $w\to \sqrt{\hat{\chi}}z$.
The solution of \eref{eq:one_body}, denoted by $x^*(z;\hat{Q},\hat{\chi})$, is given by
\begin{equation}
x^*(z;\hat{Q},\hat{\chi})=V_r(\sqrt{\hat{\chi}}z;\hat{Q}^{-1},\eta)S_r(\sqrt{\hat{\chi}}z;\hat{Q}^{-1},\eta),
\end{equation}
and is statistically equivalent to the solution of the
original problem \eref{eq:original}.
Therefore, the expected density of nonzero components in the estimate is given by 
\begin{eqnarray}
\nonumber
\hat{\rho}&\equiv\frac{1}{N}E_{\bm{y},\bm{A}}\left[\sum_{i=1}^N||\hat{\bm{x}}||_0\right]\\
&=\int Dz \mathbb{I}(|x^*(z;\hat{Q},\hat{\chi})|>0),
\end{eqnarray}
where $\mathbb{I}(x)$ is the indicator function,
which takes the value 1 when $x$ is satisfied, otherwise it is zero.
At the extremum of \eref{eq:f_replica},
the variables $Q,\chi,\hat{Q},\hat{\chi}$
are given by 
\begin{eqnarray}
\chi&=-\frac{1}{\alpha}\frac{\partial\xi(\hat{Q},\hat{\chi})}{\partial\hat{\chi}}\label{eq:RS_chi_gen}\\
Q&=\frac{1}{\alpha}\frac{\partial\xi(\hat{Q},\hat{\chi})}{\partial\hat{Q}}\label{eq:RS_Q_gen}\\
\hat{\chi}&=\frac{Q+\sigma_y^2+m_y^2}{(1+\chi)^2}\label{eq:RS_chih}\\
\hat{Q}&=\frac{1}{1+\chi}.\label{eq:RS_Qh}
\end{eqnarray}
Note that the functional form of the parameters $\hat{\chi}$
and $\hat{Q}$ does not depend on the regularization,
but the values of $\chi$ and $Q$ are regularization-dependent.

\subsection{GDF for Gaussian random predictors}

As discussed in \cite{Sakata_GDF}, the expression of GDF for Gaussian random predictors using the replica method is given by
\begin{equation}
\mathrm{df}=\frac{\chi}{1+\chi}
\label{eq:df_replica}
\end{equation}
for any regularization.
The specific form of this expression depends on the regularization.
We summarize the form of GDF for three regularizations.

\subsection{$\ell_1$ penalty}
	
For $\ell_1$ penalty,
the saddle point equation of $\chi$ is given by
\begin{equation}
\chi=\frac{\hat{\rho}}{\alpha\hat{Q}},
\label{eq:chi_L1}
\end{equation}
where $\hat{\rho}={\rm erfc}(\lambda\slash\sqrt{2\hat{\chi}})$ and 
\begin{eqnarray}
{\rm erfc}(a)=\frac{2}{\sqrt{\pi}}\int_{a}^{\infty} dze^{-z^2}.
\end{eqnarray}
Substituting \eref{eq:chi_L1} and \eref{eq:RS_Qh} into \eref{eq:df_replica},
GDF for $\ell_1$ regularization is given by
\begin{equation}
\mathrm{df}=\frac{\hat{\rho}}{\alpha}.
\label{eq:df_L1}
\end{equation}
\Eref{eq:df_L1} is the number of parameters divided by $M$, and coincides with the well-known results of GDF for LASSO \cite{GDF},
where AIC is the unbiased estimator of the prediction error.

\subsection{SCAD}

The saddle point equation of $\chi$ for SCAD regularization is given by
\begin{equation}
\chi=\frac{\hat{\rho}}{\alpha\hat{Q}}+\frac{\gamma_S}{\hat{Q}\{\hat{Q}(a-1)-1\}},
\label{eq:chi_SCAD}
\end{equation}
where 
\begin{eqnarray}
\hat{\rho}&={\rm erfc}\left(\frac{\lambda}{\sqrt{2\hat{\chi}}}\right)\\
\gamma_S&=\frac{1}{\alpha}\left(\mathrm{erfc}\left(\frac{\lambda(\hat{Q}+1)}{\sqrt{2\hat{\chi}}}\right)-\mathrm{erfc}\left(\frac{a\lambda\hat{Q}}{\sqrt{2\hat{\chi}}}\right)\right).
\end{eqnarray}
Here, $\gamma_S$ corresponds to the fraction of components 
of the regression coefficients that are 
in the transient region of SCAD (Sec.~\ref{sec:one_body_SCAD}).
\Eref{eq:chi_SCAD} leads to the following expression for GDF:
\begin{equation}
\mathrm{df}=\frac{\hat{\rho}}{\alpha}+\frac{\gamma_S}{\hat{Q}(a-1)-1}.
\label{eq:df_SCAD_replica}
\end{equation}
The coefficient of $\gamma_S$ in the second term corresponds to the rescaled variance of the estimate in the transient region.
From \eref{eq:df_SCAD_replica}, the difference between the prediction error and AIC is given by 
\begin{equation}
E_{\bm{y},\bm{A}}[\epsilon_{\mathrm{pre}}(\bm{y},\bm{A})]-E_{\bm{y},\bm{A}}[\mathrm{AIC}(\bm{y},\bm{A})]=\frac{\gamma_S\sigma_y^2}{\hat{Q}(a-1)-1}.
\label{eq:diff_SCAD}
\end{equation}
This result means that AIC underestimates the prediction error when SCAD regularization is used, and the transient region contributes to the increase in GDF.

\subsection{MCP}

The saddle point equation of $\chi$ for MCP is given by
\begin{equation}
\chi=\frac{\hat{\rho}}{\alpha\hat{Q}}+\frac{\gamma_M}{\hat{Q}(\hat{Q}a-1)},
\label{eq:chi_MCP}
\end{equation}
where
\begin{eqnarray}
\hat{\rho}&=\mathrm{erfc}\left(\frac{\lambda}{\sqrt{2\hat{\chi}}}\right),\\
\gamma_M&=\frac{1}{\alpha}\left(\mathrm{erfc}\left(\frac{\lambda}{\sqrt{2\hat{\chi}}}\right)-\mathrm{erfc}\left(\frac{a\lambda\hat{Q}}{\sqrt{2\hat{\chi}}}\right)\right),
\end{eqnarray}
and $\gamma_M$ is the fraction of the regression coefficient in the transient region of MCP
(Sec.~\ref{sec:one_body_MCP}).
\Eref{eq:chi_MCP} leads to
\begin{equation}
\mathrm{df}=\frac{\hat{\rho}}{\alpha}+\frac{\gamma_M}{\hat{Q}a-1},
\end{equation}
where the coefficient of $\gamma_M$ corresponds to the rescaled variance of the estimate in the transient region.
Hence, the difference between the prediction error and AIC is given by
\begin{equation}
E_{\bm{y},\bm{A}}[\epsilon_{\mathrm{pre}}(\bm{y},\bm{A})]-E_{\bm{y}}[\mathrm{AIC}(\bm{y},\bm{A})]=\frac{\gamma_M\sigma_y^2}{\hat{Q}a-1}.
\label{eq:diff_MCP}
\end{equation}
As with SCAD, the transient region contributes to the increase in GDF.

\subsection{Unbiasedness of the AMP-based estimator}

When the predictor matrix is homogeneous in the sense that the variance of each component is $M^{-1}$, the variable $V_\nu$ for any $\nu$ in AMP converges to $V$, which is defined by \cite{Sakata_Xu,Lenka}
\begin{equation}
V(\bm{y},\bm{A})\equiv\frac{1}{M}\sum_\mu V_\mu(\bm{y},\bm{A}).
\label{eq:V_simplify}
\end{equation}
The Gaussian random predictor considered here corresponds to this case,
and the estimator of GDF can be simplified as
\begin{eqnarray}
\hat{\mathrm{df}}^{(1^\prime)}(\bm{y},\bm{A})=\frac{V(\bm{y},\bm{A})}{1+V(\bm{y},\bm{A})}.
\label{eq:df_simplify}
\end{eqnarray}
The variable $V(\bm{y},\bm{A})$ fluctuates depending on $\bm{y}$ and $\bm{A}$, but 
its variance vanishes in the limit as $N\to\infty$ and $M\to\infty$ with $N\slash M\to\alpha$:
\begin{equation}
E_{\bm{y},\bm{A}}[(V(\bm{y},\bm{A})-E_{\bm{y},\bm{A}}[V(\bm{y},\bm{A})])^2]=0,
\end{equation}
which is a consequence of the density evolution equation \cite{Sakata_Xu,Lenka}.
Further, the density evolution equation indicates that $E_{\bm{y},\bm{A}}[V(\bm{y},\bm{A})]$ in AMP corresponds to 
$\chi$ in replica method.
Therefore, from \eref{eq:df_replica},
\begin{equation}
E_{\bm{y},\bm{A}}[\hat{\mathrm{df}}^{(1^\prime)}(\bm{y},\bm{A})]=\mathrm{df}
\label{eq:df_AMP_replica}
\end{equation}
holds for sufficiently large system size.
\Eref{eq:df_AMP_replica} means that 
the estimator $\hat{\mathrm{df}}^{(1^\prime)}(\bm{y},\bm{A})$ is asymptotically unbiased.

\begin{figure}
\centering
\includegraphics[width=3in]{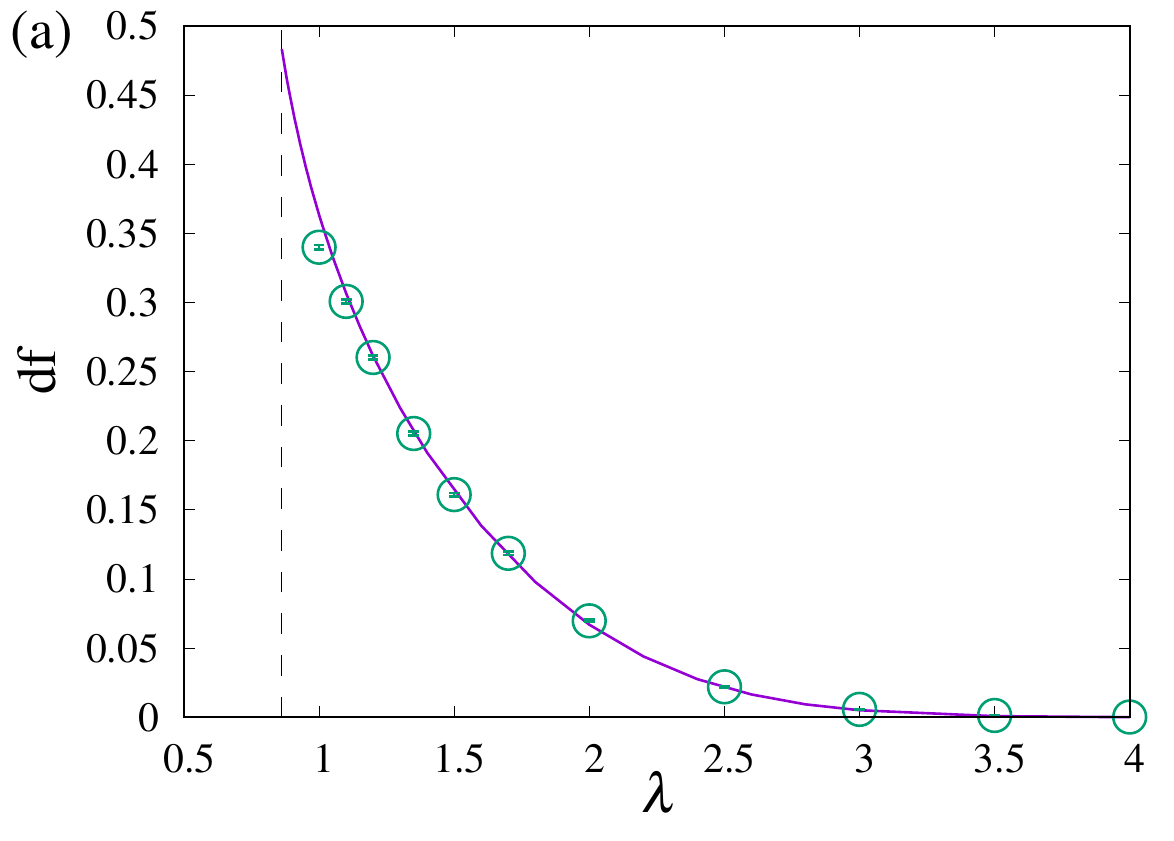}
\includegraphics[width=3in]{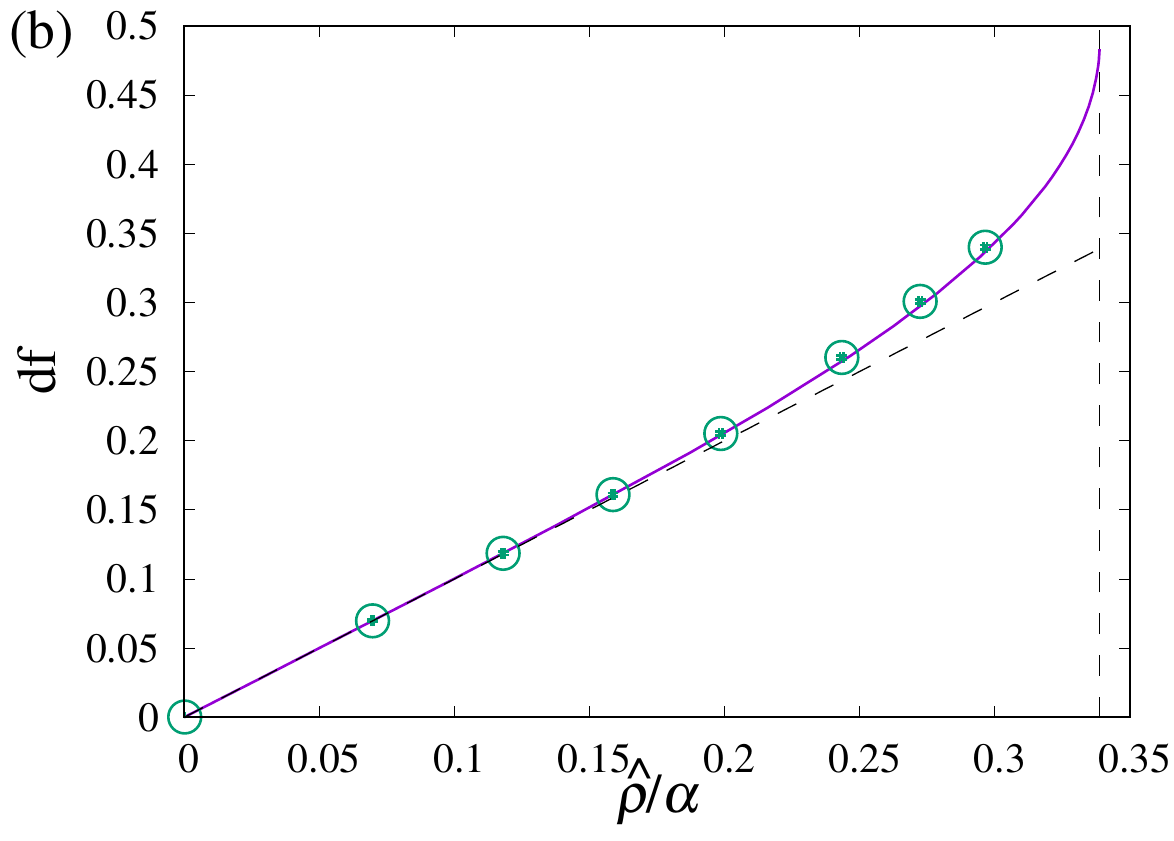}
\caption{GDF for SCAD calculated by replica method (solid line) and its estimator $\hat{\mathrm{df}}^{(1)}$ in AMP (circles) for $\alpha=0.5$, $\rho=0.1$, and $\sigma_y^2=1$:
(a) as a function of $\lambda$ and (b) as a function of $\hat{\rho}\slash\alpha$.
The dashed vertical lines represent the AT condition, and the diagonal dashed line in (b) represents the bias term of AIC.}
\label{fig:SCAD_vs_replica}
\end{figure}

\begin{figure}
\centering
\includegraphics[width=3in]{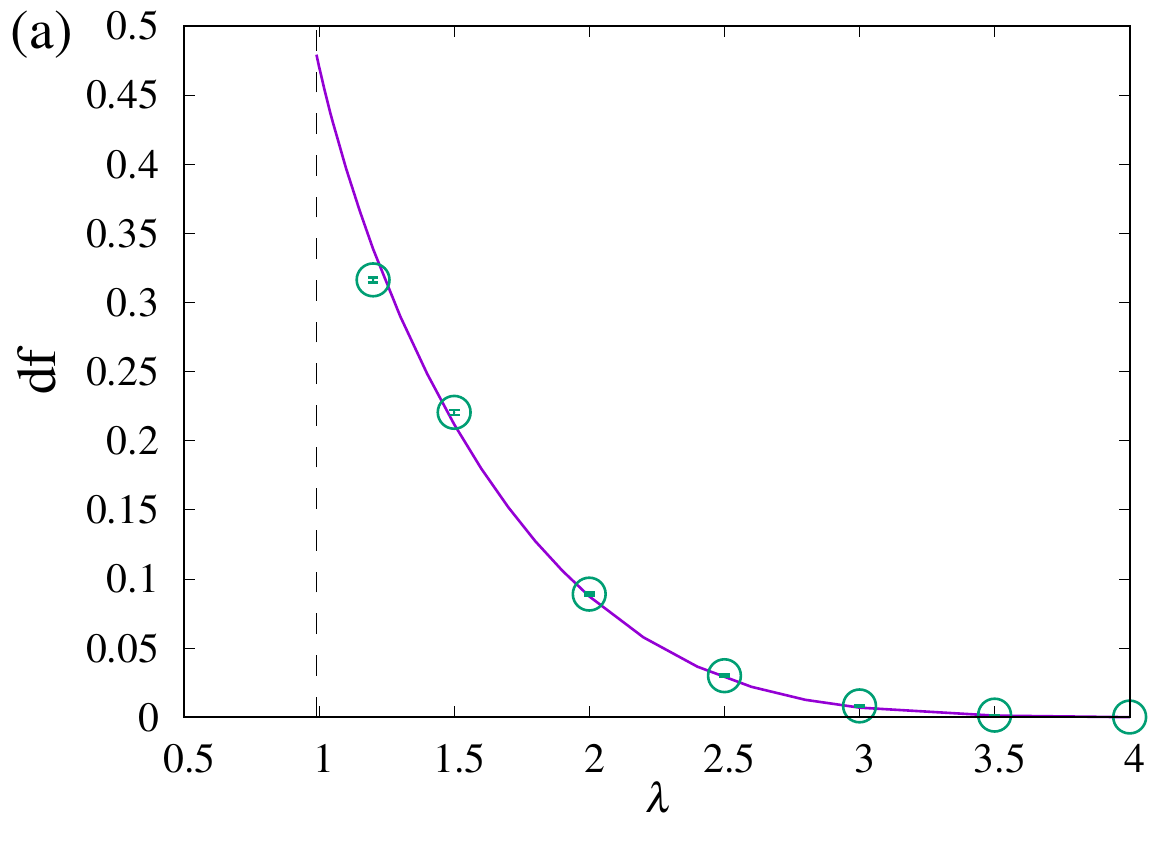}
\includegraphics[width=3in]{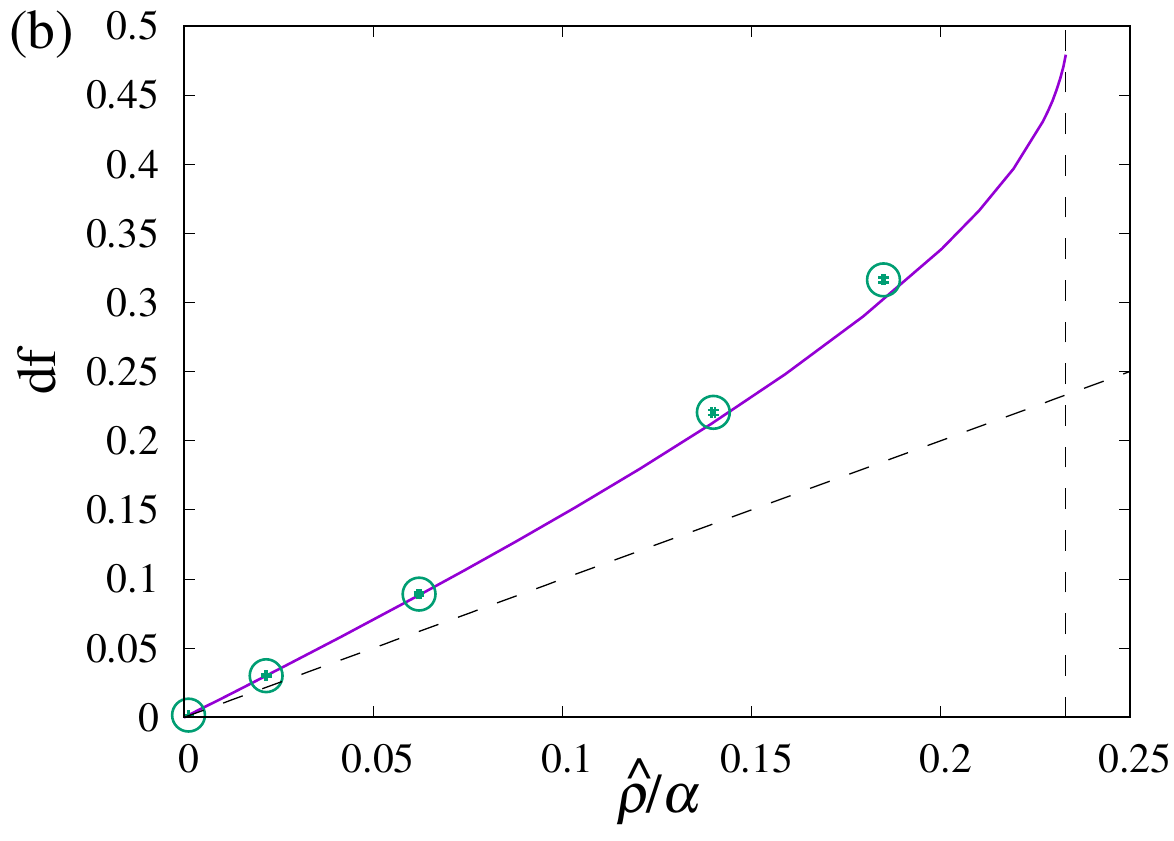}
\caption{GDF for MCP calculated by the replica method (solid line) and its estimator $\hat{\mathrm{df}}^{(1)}$ in AMP (circles) for $\alpha=0.5$, $\rho=0.1$, and $\sigma_y^2=1$:
(a) as a function of $\lambda$ and (b) as a function of $\hat{\rho}\slash\alpha$.
The dashed vertical lines represent the AT condition, and the diagonal dashed line in (b) represents the bias term of AIC.}
\label{fig:MCP_vs_replica}
\end{figure}

\Fref{fig:SCAD_vs_replica} and \Fref{fig:MCP_vs_replica} show
$\hat{\mathrm{df}}^{(1^\prime)}$ at the AMP fixed point and 
GDF calculated by the replica method with
$\alpha=0.5$ and $\sigma_y^2=1$ for SCAD and MCP, respectively.
In both figures, $a=3.7$ and (a) and (b) represent the dependence on $\lambda$ and $\hat{\rho}\slash\alpha$ of $\hat{\mathrm{df}}^{(1^\prime)}$.
In AMP, we set $N=200$, $M=100$, and the result is the average over 1000 randomly generated $\bm{y}$ and $\bm{A}$.
The vertical dashed lines represent the de Almeida--Thouless (AT) condition \cite{Sakata_Xu}.
AMP does not converge on the left of the dashed line,
and the diagonal dashed line of gradient 1 represents the bias term of AIC, $\hat{\mathrm{df}}=\hat{\rho}\slash\alpha$. 
For all parameter regions, the AMP result is consistent with the replica method,
and the difference between 
$\mathrm{df}$ and the bias term of AIC is correctly described.

\section{Application to real data}
\label{sec:application}

We applied our AMP-based estimators to the ``Communities and Crime Unnormalized Data Set'',
which is available from the UCI machine learning repository.
We used $N=70$ of the 125 predictors in the original data, and 
redefined them to reduce the similarity (correlation) between the predictors.
After preprocessing, the maximum value of the correlation between predictors was suppressed to be less than 0.685 (absolute value).
However, this was insufficient to guarantee that assumption A2 would be satisfied.
In addition, a part of the coefficients had larger values for which $O(1\slash\sqrt{M})$ did not hold, 
and so assumption A1 was also violated.
Among the 2215 response variables in the original data, we used $M=302$ response variables for which the corresponding row vectors in the predictor matrix were not missing.
The original data contained 18 kinds of response variables.
Here, we demonstrate the model selection based on the proposed estimator using the response variable
of ``the number of murders per 100K population''.

The simulation condition was as follows. The predictors and output variables were standardized.
First, we computed the OLS estimate
$\hat{\bm{x}}_{\mathrm{OLS}}=\bm{A}^{+}\bm{y}$, where $\bm{A}^+$ is the pseudo-inverse matrix of $\bm{A}$.
The signal $\bm{x}^0$
estimated in this simulation was generated from the OLS estimate.
We denote the set of indices of the largest $K$ components of $\hat{\bm{x}}^{\mathrm{OLS}}$ (in absolute value)
as $S_K$, and set $x_i^0=\hat{x}_i^{\mathrm{OLS}}$ if $i\in S_K$ and $x_i^0=0$ otherwise.
We define $\hat{\sigma}_y^2=M^{-1}||\bm{y}-\bm{A}\bm{x}^0||_2^2$ and 
consider a synthetic model
\begin{equation}
\bm{y}^*=\bm{A}\bm{x}^0+\hat{\sigma_y}\bm{\xi},
\label{eq:crime_synthetic}
\end{equation}
where each component of $\bm{\xi}\in\mathbb{R}^M$
is an i.i.d. Gaussian random variable with mean 0 and variance 1.

\begin{figure}
\centering
\includegraphics[width=3in]{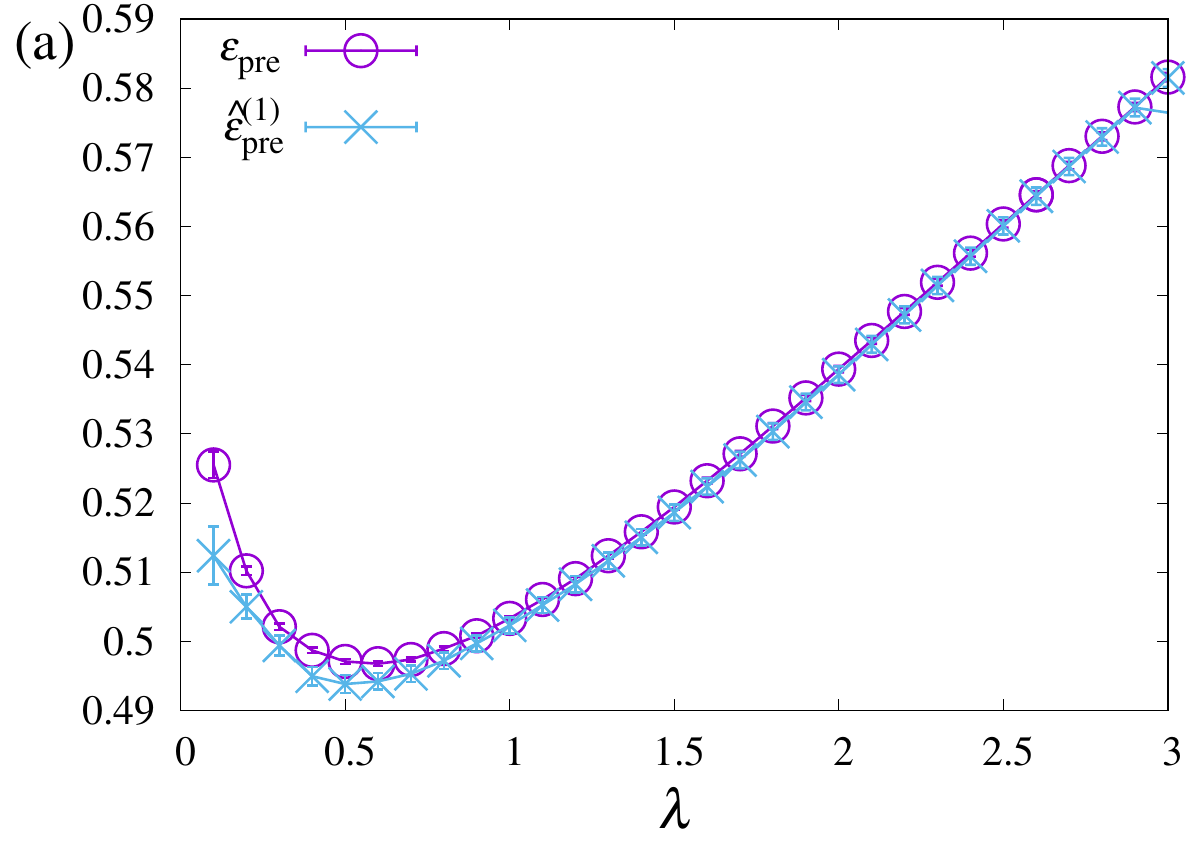}
\includegraphics[width=3in]{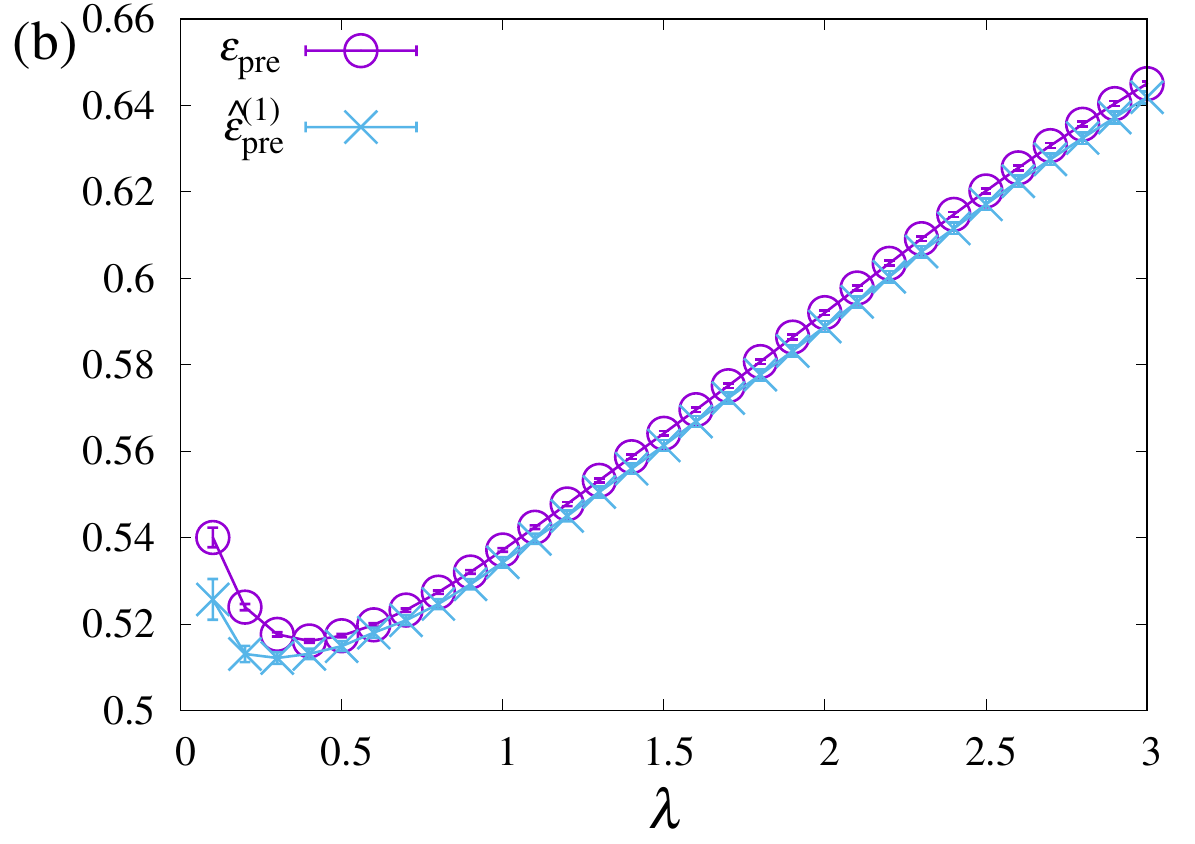}
\caption{$\lambda$-dependence of the prediction error and its estimator for ``Communities and Crime Unnormalized Data Set'' under $\ell_1$ penalty with (a) $K=7$ and (b) $K=14$.}
\label{fig:L1_crime}
\end{figure}

\Fref{fig:L1_crime} shows the $\lambda$-dependence of the prediction error and our proposed estimator $\hat{\epsilon}_{\mathrm{pre}}^{(1)}$ under the $\ell_1$ penalty for $K=7$ and $K=14$.
To evaluate the prediction error,
we prepared 1000 test samples and imitated the expectation according to the generative process using the sample average.
The value of the estimator was averaged over 1000 training samples to describe its typical behavior, and the averaged value was compared with the prediction error.
In the case of the $\ell_1$ penalty, the behavior of $\hat{\epsilon}_{\mathrm{pre}}^{(1)}$ is similar to the prediction error, and it can select the model that minimizes the prediction error.

However, in the case of SCAD and MCP, 
the proposed estimators do not provide such a good match to the prediction error.
\Fref{fig:crime_a3.7} shows the $\lambda$-dependence of the prediction error,
the estimators of prediction error, and AIC at $K=7$ for SCAD and MCP.
The regularization parameter $a$ is set to 3.7 in both cases.
The prediction error attains a minimum at $\lambda=0.8$ for SCAD and $\lambda=0.9$ for MCP, but AIC cannot detect these minima.
AIC tends to deviate from the prediction error as $\lambda$ decreases, and reaches a minimum at smaller values of $\lambda$. In other words, the model selected based on AIC has more non-zero components than that selected by the minimization of the prediction error.
In both cases, 
$\hat{\epsilon}_{\mathrm{pre}}^{(1)}$ is slightly larger than AIC.
The difference between AIC and $\hat{\epsilon}_{\mathrm{pre}}^{(1)}$ is interpreted as the contribution of the estimates in the transient region, referring to \eref{eq:diff_SCAD} and \eref{eq:diff_MCP} for a Gaussian random predictor and response.
However, under assumptions A1 and A2, the AMP-based estimator of GDF underestimates its true value. We introduce a heuristic method to modify the proposed estimator in the following section.

\section{Correction of the estimator taking into account correlation between predictors}
\label{sec:correction}

\begin{figure}
\centering
\includegraphics[width=3in]{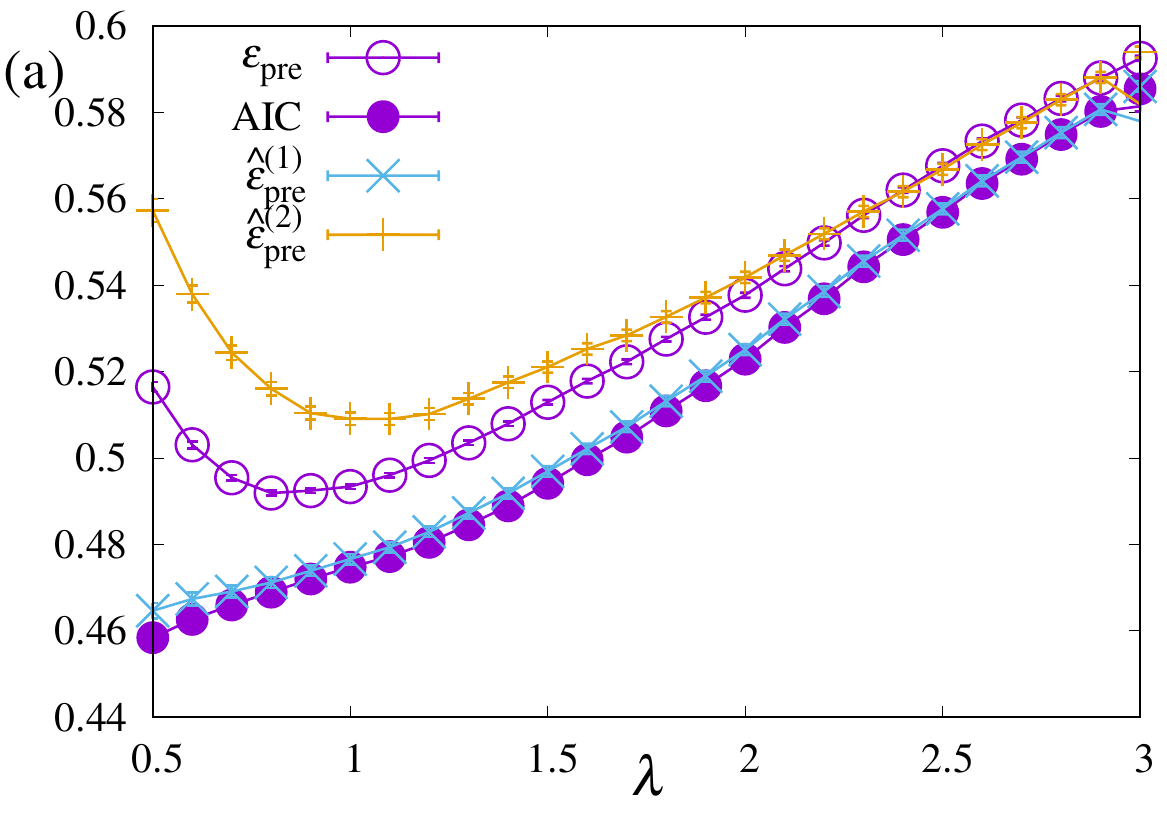}
\includegraphics[width=3in]{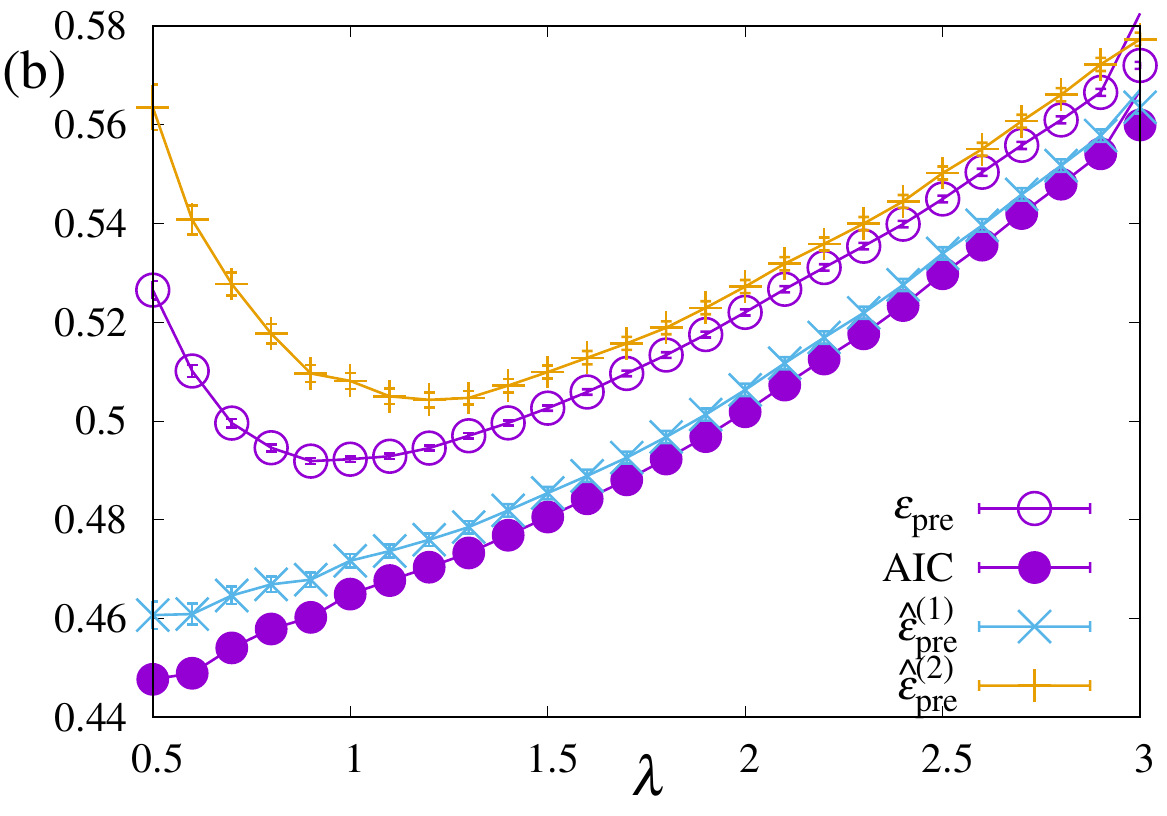}
\caption{$\lambda$-dependence of the prediction error ($\circ$), AIC ($\bullet$) under (a) SCAD and (b) MCP for ``Communities and Crime Unnormalized Data Set''.}
\label{fig:crime_a3.7}
\end{figure}

The estimator \eref{eq:df_AMP_1} is an increasing function of $V_\mu (>0)$, and hence
the underestimation of the prediction error is caused by that of $V_\mu$.
We correct the value of the rescaled variance $\{v_i\}$ included in $V_\mu$
by considering the off-diagonal elements of $\bm{A}^{\mathrm{T}}\bm{A}$,
after the convergence of AMP.

From \eref{eq:AMP_v_i} and \eref{eq:f_c_and_f_a},
the variable $v_i$ corresponds to the variation of $x_i$ with respect to
$R_i\slash\Sigma^2$.
As shown in \eref{eq:f_a_one_body},
$R_i\slash\Sigma^2$ is the effective field of the single-body minimization problem of $x_i$.
The assumptions introduced for the derivation of AMP 
convert the original problem to the single-body problem,
ignoring the correlation between the predictors.
We go back to the original problem
with respect to the support set at the fixed point of AMP,
to re-calculate $\bm{v}$ and $\bm{V}$ taking into account the correlation between the predictors.
The corrected $\bm{v}$ and $\bm{V}$ are denoted by $\tilde{\bm{v}}$ and $\tilde{\bm{V}}$, respectively.

We denote the support indices at the 
fixed point of AMP by $K$,
and denote the vector of the variables $x_j$ ($j\in K$) as $\bm{x}_K\in\mathbb{R}^{|K|}$,
where $|K|$ is the size of the support set $K$ and  
the $i$-th component of $K$ is denoted by $K_i$.
Furthermore, the submatrix of $\bm{A}$ consists of $i(\in K)$-th columns and
is denoted by $\bm{A}_K$.
We consider the original minimization problem with respect to $\bm{x}_K$ under an infinitesimal external field $\bm{h}\in\mathbb{R}^{|K|}$ as \cite{GDF,CV_Obuchi}
\begin{equation}
\min_{\bm{x}_K}\left\{\frac{1}{2}||\bm{y}-\bm{A}_K\bm{x}_K||_2^2+\bm{J}(\bm{x}_K)-\bm{h}^{\mathrm{T}}\bm{x}_K\right\}.
\end{equation}
The minimizer is the solution of the following equation
\begin{equation}
-\bm{A}_K^{\mathrm{T}}\bm{y}+(\bm{A}_K^{\mathrm{T}}\bm{A}_K)\bm{x}_K+\bm{J}^\prime-\bm{h}=0,
\label{eq:x_support}
\end{equation}
where the $i$-th component of $\bm{J}^\prime\in\mathbb{R}^{|K|}$,
denoted by $J_i^\prime$, is given by
\begin{equation}
J^\prime_{i} = \frac{\partial J(x_{K_i})}{\partial x_{K_i}}.
\end{equation}
We correct $\bm{v}$
by implementing the variation with respect to the external field in the original problem. 
The infinitesimal external field helps the procedure as
\begin{equation}
\tilde{v}_{K_i}=\frac{\partial x_{K_i}}{\partial h_{K_i}}\Big|_{\bm{h}\to\bm{0}}.
\label{eq:v_derivative}
\end{equation}
Therefore, the corrected variable $\tilde{\bm{V}}$ is expressed as
\begin{equation}
\tilde{V}_\mu=\sum_{i\in K} A_{\mu i}^2\tilde{v}_{i}.
\label{eq:V_tilde}
\end{equation}
Using $\tilde{V}_\mu$, we define the estimator of GDF as
\begin{equation}
\hat{\mathrm{df}}^{(2)}(\bm{y})=\frac{1}{M}\sum_{\nu=1}^M\frac{\tilde{V}_\nu(\bm{y})}{1+\tilde{V}_\nu(\bm{y})},
\label{eq:df_2}
\end{equation}
and the corresponding estimator of the prediction error as
\begin{equation}
\hat{\epsilon}_{\mathrm{pre}}^{(2)}(\bm{y})\equiv\epsilon_{\mathrm{train}}(\bm{y})+2\sigma_y^2\hat{\mathrm{df}}^{(2)}(\bm{y}).
\label{eq:estimator_pre_err}
\end{equation}

\Fref{fig:pseudocode} is the pseudocode for the
calculation of the AMP-based estimator of the prediction error.
As a by-product of this correction, $\tilde{v}_i$ can be calculated using other algorithms
that do not update $v_i$,
such as coordinate descent (CD) and Alternating Direction Method of Multipliers (ADMM).
It is known that the convergence of AMP is slow when the predictor matrix is not i.i.d. Gaussian. In such cases, the combination of the AMP-based estimator with CD or ADMM is useful, as these techniques are less influenced by the properties of the predictor matrix than AMP.

\begin{figure}
\renewcommand{\thepseudocode}{\arabic{pseudocode}}
\setcounter{pseudocode}{0}
\begin{pseudocode}[ruled]
{AMP-based estimator of the prediction error}
{\hat{\epsilon}_{\mathrm{pre}}^{(2)}(\bm{y})}

1)\ \mbox{\bf Get the fixed point of AMP}:\\
\hspace{1.0cm}\mbox{Update variables according to 
\eref{eq:AMP_V}-\eref{eq:AMP_v_i} until they converge.}\\
\hspace{1.0cm}\mbox{Calculate the training error $\epsilon_{\mathrm{train}}(\bm{y})$ at the fixed point.}\\

2)\ \mbox{\bf Find $K$}: \\
\hspace{1.0cm}\mbox{Set the support of $\bm{a}$ at the fixed point of AMP as $K$, }\\
\hspace{1.0cm}\mbox{and 
construct a submatrix $\bm{A}_K$.} \\

3)\ \mbox{\bf Get $\bm{x}_K$}:\\
\hspace{1.0cm}\mbox{Solve \eref{eq:x_support}
and set the solution as $\bm{x}_K$.}\\

4)\ \mbox{\bf Calculate $\tilde{\bm{v}}$}:\\
\hspace{1.0cm}\mbox{Calculate \eref{eq:v_derivative} for all components in the support.}\\

5)\ \mbox{\bf Get estimator of df}:\\
 \hspace{1.0cm}
 \mbox{Calculate $\tilde{\bm{V}}$ as \eref{eq:V_tilde} and put it into \eref{eq:df_2},
 then obtain $\hat{\mathrm{df}}^{(2)}(\bm{y})$}.\\

6)\ \mbox{\bf Get estimator of the prediction error}:\\
 \hspace{1.0cm}
 \mbox{According to \eref{eq:estimator_pre_err},
 calculate the estimator of the prediction error $\hat{\epsilon}_{\mathrm{pre}}^{(2)}(\bm{y})$.}
 
\end{pseudocode}
\protect
\caption{Pseudocode for the calculation of AMP-based estimator of the prediction error, $\hat{\epsilon}_{\mathrm{pre}}^{(2)}(\bm{y})$.}
\label{fig:pseudocode}
\end{figure}

%

We summarize the specific form of $\tilde{v}_i$ for SCAD and MCP.
In the case of SCAD,
the vector $\bm{J}^\prime$ is given by
\begin{equation}
\bm{J}^\prime=\lambda\bm{\Psi}\mathrm{sgn}(\bm{x}_K)-\bm{\Phi}\left(\frac{\bm{x}_K-a\lambda\mathrm{sgn}(\bm{x}_K)}{a-1}\right),
\label{eq:J_dif_SCAD}
\end{equation}
where $\bm{\Psi}\in\mathbb{R}^{|K|\times |K|}$ and $\bm{\Phi}\in\mathbb{R}^{|K|\times |K|}$ are diagonal matrices whose components are given by
\begin{eqnarray}
{\Psi}_{i,i}&=\mathbb{I}(|x_{K_i}|\leq\lambda)\\
{\Phi}_{i,i}&=\mathbb{I}(\lambda<|x_{K_i}|\leq a\lambda).
\end{eqnarray}
The solution of \eref{eq:x_support} is given by 
\begin{equation}
\bm{x}_K=(\bm{A}_K^{\mathrm{T}}\bm{A}_K-(a-1)^{-1}\bm{\Phi})^{-1}(\bm{A}_K^{\mathrm{T}}\bm{y}+\bm{h}-\lambda\mathrm{sgn}(\bm{x}_K)),
\end{equation}
and setting $\bm{U}=(\bm{A}_K^{\mathrm{T}}\bm{A}_K-(a-1)^{-1}\bm{\Phi})^{-1}$
gives the rescaled variance 
\begin{equation}
\tilde{v}_{K_i}=U_{ii}.
\end{equation}

In the case of MCP, the vector $\bm{J}^\prime$ is given by 
\begin{equation}
\bm{J}^\prime=\bm{\Phi}\left(\lambda\mathrm{sgn}(\bm{x}_K)-\frac{\bm{x}_K}{a}\right),
\end{equation}
where $\bm{\Phi}$ is the diagonal matrix whose components are 
\begin{equation}
{\Phi}_{i,i}=\mathbb{I}(|x_{K_i}|\leq a\lambda).
\end{equation}
Setting $\bm{U}=(\bm{A}_K^{\mathrm{T}}\bm{A}_K-a^{-1}\bm{\Phi})^{-1}$,
we obtain
\begin{equation}
\tilde{v}_{K_i}=U_{ii}.
\end{equation}

\begin{figure}
\centering
\includegraphics[width=3in]{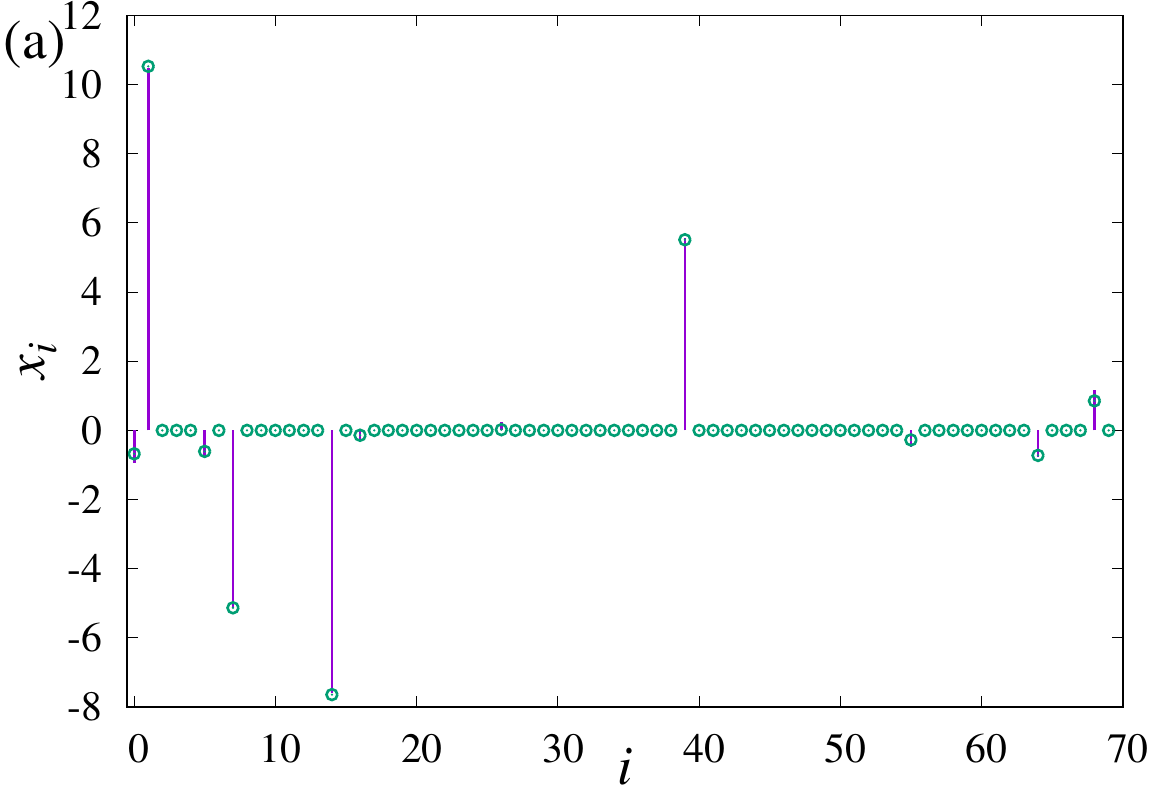}
\includegraphics[width=3in]{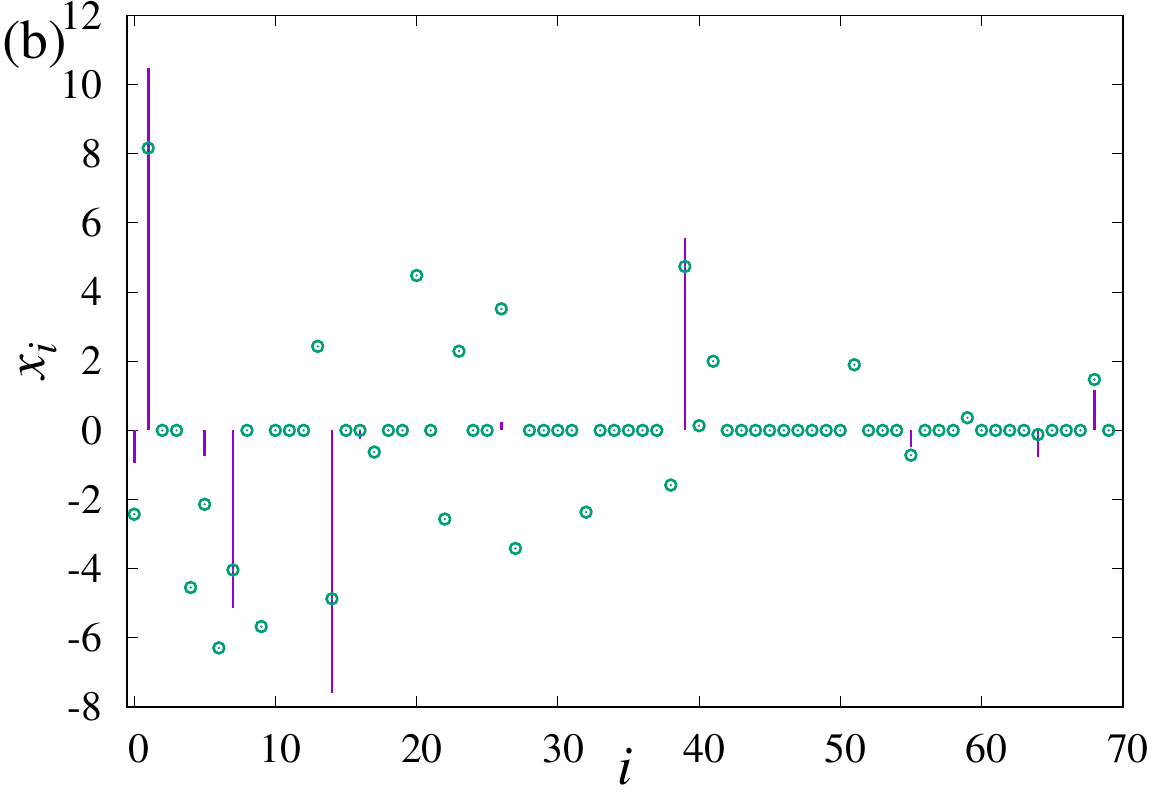}
\caption{Comparison of models for SCAD.
Models with the minimum prediction error are denoted by bars, and the circles in
(a) and (b) represent the model given by the minimum of $\hat{\epsilon}_{\mathrm{pre}}^{(2)}$ and AIC, respectively.
The minima of the prediction error, $\hat{\epsilon}_{\mathrm{pre}}^{(2)}$,
and AIC are obtained with $\lambda=0.8,~a=3.7$, $\lambda=1,~a=3.2$, and
$\lambda =0.45,~a=3.7$, respectively.}
\label{fig:model_comparison_SCAD}
\end{figure}

\begin{figure}
\centering
\includegraphics[width=3in]{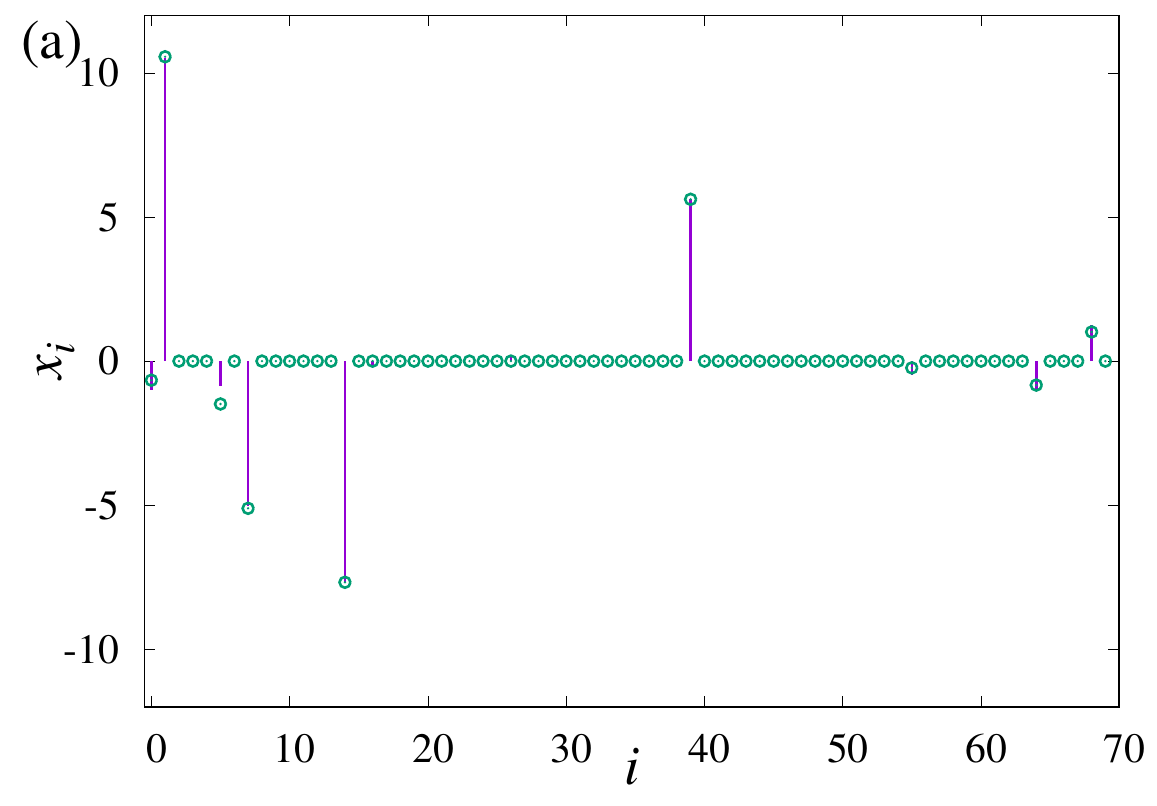}
\includegraphics[width=3in]{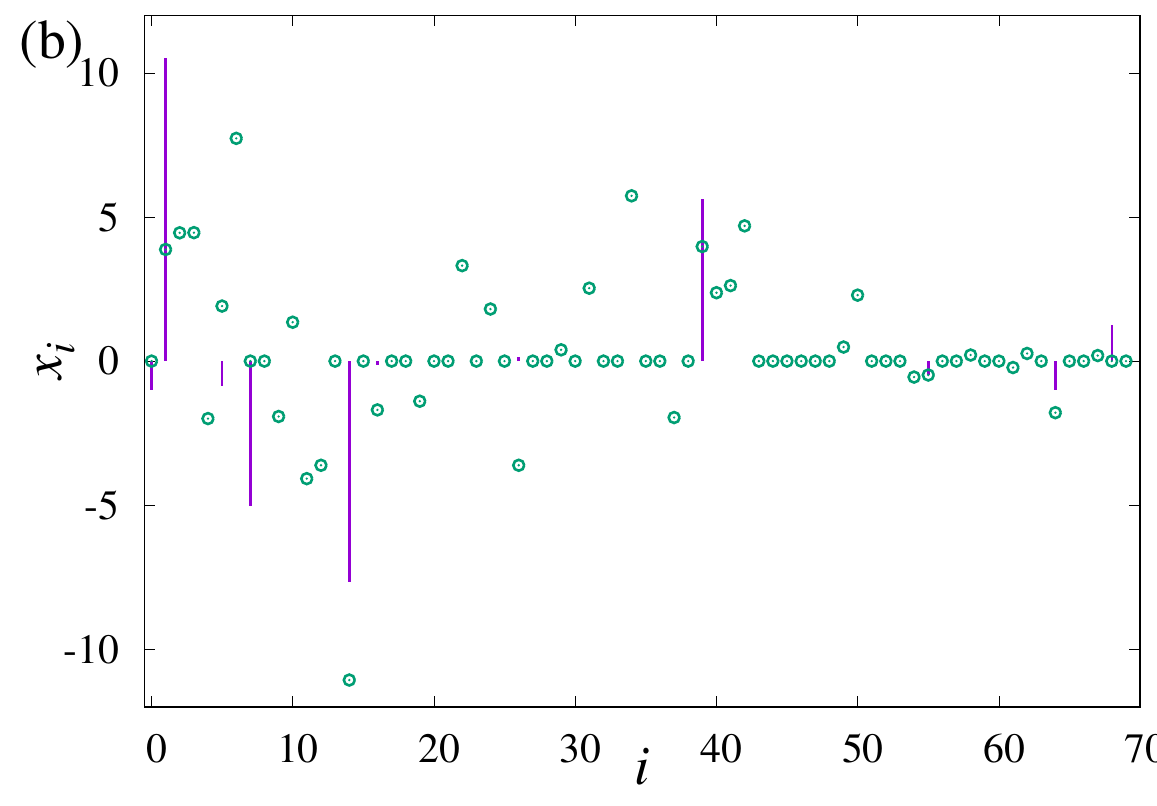}
\caption{Comparison of models for MCP. 
Models with the minimum prediction error are denoted by bars, and the circles in
(a) and (b) represent the model given by the minimum of $\hat{\epsilon}_{\mathrm{pre}}^{(2)}$ and AIC, respectively.
The minima of the prediction error, $\hat{\epsilon}_{\mathrm{pre}}^{(2)}$,
and AIC are obtained with $\lambda=0.9,~a=3.7$, $\lambda=1.2,~a=3.6$, and
$\lambda =0.45,~a=3.7$, respectively.}
\label{fig:model_comparison_MCP} 
\end{figure}

We examined the performance of the modified AMP-based estimator $\hat{\epsilon}_{\mathrm{pre}}^{(2)}$ in terms of the prediction error for the synthetic model \eref{eq:crime_synthetic}
under SCAD and MCP. The results are shown in \Fref{fig:crime_a3.7}.
The estimator $\hat{\epsilon}_{\mathrm{pre}}^{(2)}$ captures the qualitative behavior of the prediction error
better than AIC and $\hat{\epsilon}_{\mathrm{pre}}^{(1)}$.
In particular, $\hat{\epsilon}_{\mathrm{pre}}^{(2)}$ attains a minimum at a value of $\lambda$ that is 
close to that which minimizes the prediction error. Thus,
it can effectively emulate the prediction error in the model selection process.
Figures \ref{fig:model_comparison_SCAD} and \ref{fig:model_comparison_MCP}
show the models selected according to the minimum prediction error
given by $\hat{\epsilon}_{\mathrm{pre}}^{(2)}$ and AIC
for SCAD and MCP, respectively.
The model selected based on the minimization of $\hat{\epsilon}_{\mathrm{pre}}^{(2)}$
is closer to that selected by the prediction error than the model given by AIC.
These results indicate the advantage of the estimator $\hat{\epsilon}_{\mathrm{pre}}^{(2)}$ over AIC
when it is necessary to select a model based on the minimization of the prediction error
under nonconvex penalties. 

Matlab codes for calculating the AMP-based estimator and
the data used in this simulation are provided on our webpage.

\section{Summary and Conclusion}
\label{sec:summary}

We have proposed an AMP-based estimator of GDF and the prediction error
for penalized linear regression problems.
The proposed GDF estimator is given by the variable $V_\mu$ in AMP for any regularization.
The asymptotic property of the estimator $\hat{\mathrm{df}}^{(1)}$ for a Gaussian random predictor and response variable is consistent with the result derived by the replica method,
and the asymptotic unbiasedness is mathematically guaranteed.
The prediction error estimator using $\hat{\mathrm{df}}^{(1)}$ could be improved when applied to real data whose predictors are correlated.
We corrected the estimator by considering the correlation between predictors, and
thus constructed $\hat{\mathrm{df}}^{(2)}$. 
We demonstrated the model selection process using the proposed estimators for nonconvex sparse penalties with real data, where AIC does not correspond to the unbiased estimator of the prediction error.
The proposed estimator selects models that are close to the model that minimizes the prediction error.

Our proposed method can determine the regularization parameter, particularly for regularizations in which the unbiased estimator of the prediction error cannot be analytically derived.
However, the correspondence between the proposed estimator and the prediction error is not mathematically guaranteed, except when the predictor matrix is i.i.d. Gaussian.
The validity of our proposed method should be examined under several scenarios,
especially for nonconvex penalties.
For the real data used in this paper, qualitatively similar results are obtained for other response variables that the model selected based on our proposed estimators is close to the minimum of the prediction error.
The generalization of our discussion to correlated predictors and rank-deficient predictors is required 
for more mathematically rigorous discussions \cite{SAMP}.
Furthermore, in practical usage, an efficient method of finding the minimum of the estimated prediction error should be developed.

\ack

This work is supported by JSPS KAKENHI (No.16K16131) and Shiseido Female Researcher Science Grant.

\end{document}